\documentclass{article}

\PassOptionsToPackage{numbers, sort&compress}{natbib}
\usepackage[final, main]{neurips_2026}

\usepackage[utf8]{inputenc}
\usepackage[T1]{fontenc}
\usepackage{amsmath,amssymb,amsthm}
\usepackage{amsfonts}
\usepackage{bm}
\usepackage{graphicx}
\usepackage{booktabs}
\usepackage{array}
\usepackage{nicefrac}
\usepackage{microtype}
\usepackage{xcolor}
\usepackage{url}
\usepackage{tikz}
\usetikzlibrary{shapes.geometric,arrows.meta,positioning,fit,backgrounds,calc}
\usepackage{pgfplots}
\pgfplotsset{compat=1.18}
\usepgfplotslibrary{groupplots,colormaps}
\usetikzlibrary{arrows.meta,positioning,calc,shapes.misc}
\definecolor{cGray}{HTML}{6B7280}
\definecolor{cBlue1}{HTML}{60A5FA}
\definecolor{cBlue2}{HTML}{1E3A8A}
\definecolor{cK}{HTML}{9CA3AF}
\definecolor{cGrid}{HTML}{E3E3E3}
\definecolor{cOrange}{HTML}{CC6600}
\definecolor{cOrangeFace}{HTML}{FCE5CD}
\definecolor{cBlueFace}{HTML}{CFE2F3}
\definecolor{cBlueEdge}{HTML}{1F4E79}
\definecolor{cRed}{HTML}{D62728}
\definecolor{cBlueS}{HTML}{1F77B4}
\pgfplotsset{
  crstyle/.style={
    axis line style={line width=0.5pt, draw=black!80},
    axis x line*=bottom, axis y line*=left,
    tick style={line width=0.4pt, draw=black!80},
    ymajorgrids, grid style={cGrid, line width=0.4pt},
    tick label style={font=\scriptsize},
    label style={font=\scriptsize},
    title style={font=\footnotesize, align=left, anchor=south west, at={(0,1)}, yshift=1pt},
    legend style={draw=none, fill=none, font=\scriptsize},
    legend cell align=left,
    every axis plot/.append style={line width=0.8pt},
  }
}

\colorlet{cBase}{cGray}
\colorlet{cHead}{cBlue1}
\colorlet{cStrong}{cBlue2}
\colorlet{cPerm}{cOrange}
\colorlet{cRef}{black!55}
\colorlet{cLam1}{cBlue1!55!white}
\colorlet{cLam2}{cBlue1}
\colorlet{cLam3}{cBlue2!70!cBlue1}
\colorlet{cLam4}{cBlue2}
\colorlet{cLam5}{cBlue2!60!black}
\pgfplotsset{
  crfig/.style={crstyle,
    tick label style={font=\footnotesize}, label style={font=\footnotesize},
    legend style={font=\footnotesize, draw=none, fill=white, fill opacity=0.9, text opacity=1, inner sep=1.5pt, row sep=-2pt},
    title style={font=\small, align=left, anchor=south west, at={(0,1)}, yshift=1pt},
    grid style={cGrid, line width=0.5pt},
    every axis plot/.append style={line width=1.0pt},
  },
  crbar/.style={crfig, ybar, bar shift=0pt, xtick style={draw=none}, enlarge x limits=0.22,
    nodes near coords, nodes near coords style={font=\footnotesize, /pgf/number format/fixed}},
}
\tikzset{lReal/.style={solid}, lPerm/.style={dashed}, lRef/.style={densely dotted, cRef, line width=0.6pt}}

\usepackage[pagebackref,breaklinks,colorlinks]{hyperref}
\usepackage[capitalize]{cleveref}
\crefname{figure}{Figure}{Figures}
\usepackage{sidecap}
\renewcommand*{\backref}[1]{}
\renewcommand*{\backrefalt}[4]{%
  \ifcase #1 (Not cited.)%
  \or {\footnotesize\textcolor{gray!70!black}{$\hookrightarrow$ p.~#2}}%
  \else {\footnotesize\textcolor{gray!70!black}{$\hookrightarrow$ pp.~#2}}%
  \fi}

\graphicspath{{figures/}}

\newcommand{\Lcon}{\mathcal{L}_{\mathrm{con}}}
\newcommand{\LOU}{\mathcal{L}_{\mathrm{OU}}}

\newcommand{\bbeta}{\bm{\beta}}
\newcommand{\dphylo}{d_{\mathrm{tax}}}

\newcommand{\alphainf}{\alpha\!\to\!\infty}

\newcommand{\OUcov}[2]{%
  \frac{\sigma^{2}}{2\alpha}\exp\!\big(-\alpha\,\dphylo(#1,#2)\big)}

\newtheorem{proposition}{Proposition}
\newtheorem{remark}{Remark}

\title{Beyond Family Labels: A Taxonomic Ornstein--Uhlenbeck Prior
for Avian 3D Shape Recovery}

\makeatletter
\@ifundefined{iflatexml}{\newif\iflatexml\latexmlfalse}{}
\makeatother

\iflatexml
\author{%
  Yunchan Jeon \\
  Department of Computer Science and Engineering\\
  Chung-Ang University, South Korea \\
  \texttt{ycj00@cau.ac.kr}
  \and
  SooGon Kim \\
  Department of Computer Science and Engineering\\
  Chung-Ang University, South Korea \\
  \texttt{gon1397@cau.ac.kr}
  \and
  Seungwook Kim \\
  GenGenAI, South Korea \\
  \texttt{wookiekim@gengen.ai}
  \and
  CheolWon Lee \\
  Department of Computer Engineering\\
  Konkuk University, South Korea \\
  \texttt{cheolwonlee@kku.ac.kr}
  \and
  Jongmin Lee \\
  Department of Computer Science and Engineering\\
  Chung-Ang University, South Korea \\
  \texttt{jmlee@cau.ac.kr}
}
\else
\author{%
  Yunchan Jeon$^{1,*}$ \hspace{0.45em}
  SooGon Kim$^{1,*}$ \hspace{0.45em}
  Seungwook Kim$^{2}$ \hspace{0.45em}
  CheolWon Lee$^{3}$ \hspace{0.45em}
  Jongmin Lee$^{1,\dagger}$ \\[0.25cm]
  $^{1}$Department of Computer Science and Engineering, Chung-Ang University, South Korea \\
  $^{2}$GenGenAI, South Korea \\
  $^{3}$Department of Computer Engineering, Konkuk University, South Korea \\[0.15cm]
  {\footnotesize\texttt{\{ycj00, gon1397, jmlee\}@cau.ac.kr} \hspace{0.7em}
  \texttt{wookiekim@gengen.ai} \hspace{0.7em}
  \texttt{cheolwonlee@kku.ac.kr}} \\[0.15cm]
  $^{*}$Equal contribution \qquad
  $^{\dagger}$Corresponding author
}
\fi

\begin{document}

\maketitle

\begin{abstract}
Recovering 3D pose and shape of birds from single images is bottlenecked by
the long tail of avian diversity: out of $\sim$$10{,}000$ species, only a
few dozen have 3D-labeled training data, and in-the-wild test sets routinely
contain species never seen at training. Generalizing across clades is
therefore a central problem. The current state of the art,
AniMer+~\citep{animerplus2025}, mitigates it with a
family-level contrastive loss that conveys only a binary same-family
indicator per pair and discards the continuous structure of the species
tree.
We propose a \textbf{taxonomic Ornstein--Uhlenbeck prior} on the shape
code of the AVES parametric bird mesh model: a multivariate Gaussian whose covariance decays exponentially with
Linnaean rank distance, with a learnable selection strength and noise scale. The prior
generalizes family contrastive into a continuous, generative form, exposing
the graded cross- and within-family similarity that a binary indicator cannot,
and brings continuous comparative phylogenetics in as a biology-grounded
inductive bias for the long tail.
Our method improves over AniMer+ on both metrics of the unseen-species
CowBird benchmark and on CtrlAVES3D, while remaining comparable on
seen-species CUB. A permutation ablation isolates taxonomic structure
from generic Gaussian smoothing, and a rank-granularity ablation identifies
cross-order distinctions (the structure beyond a family-vs-not indicator)
as the dominant source of the unseen-species gain.
\end{abstract}

\section{Introduction}
\label{sec:introduction}

\begin{figure}[t]
\centering
\scalebox{0.95}{\includegraphics[width=\linewidth]{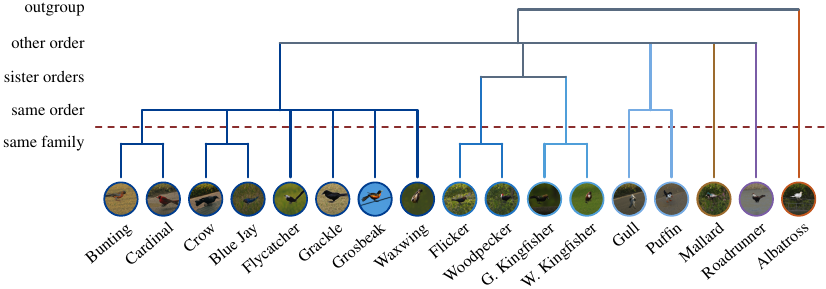}}
\vspace{-0.2cm}
\caption{\textbf{Beyond family labels.}
  The $17$ training species of CtrlAVES3D on the Linnaean
  tree~\citep{ioc-bird-list}, colored by order. The height at which two
  lineages join marks their lowest shared rank, from same family up to
  outgroup; each rank sets the distance $d_{\mathrm{tax}}$ that our
  prior reads (\cref{fig:architecture}b).
  The family-supervised contrastive loss of
  AniMer+~\citep{animerplus2025} keeps a single bit of this tree: whether
  a pair joins below the dashed red line.
  Our Ornstein--Uhlenbeck (OU) prior uses every rank, pulling the shape
  codes of two species together more strongly the more closely they are
  related, so the shape space carries graded structure rather than a
  same-family indicator.}
\label{fig:teaser}
\end{figure}

\paragraph{Motivation: the long-tail problem in avian 3D recovery.}
Single-image articulated 3D reconstruction has progressed rapidly for
animals, from species-agnostic fits of the Skinned Multi-Animal Linear
(SMAL) model~\citep{smal2017} to breed-aware
priors~\citep{barc2022}, web-scale category models~\citep{3dfauna2024}, and
most recently to family-aware mixture-of-experts transformers for joint
mammal--bird recovery~\citep{animerplus2025}. Birds, however, sharpen the
classical long-tail problem to an extreme. The world hosts on the order of
$10\,000$ extant bird species, yet 3D-labeled training data exist for only
a few dozen, and important in-the-wild benchmarks
(e.g.\ CowBird~\citep{badger2020}) contain species \emph{never seen} at
training. A practical estimator must therefore
generalize across species: not just to new individuals or new poses, but to
new clades (groups of species sharing a common ancestor) whose body
plans the model has never been supervised on.
Bridging this gap requires a regularizer that injects \emph{prior
knowledge} about how shape varies across species, knowledge that exists
abundantly in evolutionary biology but is not currently exploited by
single-view recovery pipelines.

\paragraph{Background problem: family supervised contrastive discards
continuous taxonomic structure.}
The current bird state of the art, AniMer+~\citep{animerplus2025},
addresses the long tail with a family-level supervised-contrastive
(SupCon)~\citep{supcon2020} regularizer $\mathcal{L}_{\mathrm{con}}$
on the latent shape code. By construction $\mathcal{L}_{\mathrm{con}}$
ingests only a binary same-family indicator per pair and discards
\emph{all} taxonomic structure below and above the family cut. Two species in the same family are
pushed equally close regardless of whether they diverged $5$ or
$50$ million years ago; two species in different families are pushed equally far
regardless of whether they share an order or sit on opposite ends of
Aves. Yet decades of phylogenetic comparative methods in evolutionary
biology~\citep{hansen1997,butler2004} show that continuous traits
across species are well modeled by an Ornstein--Uhlenbeck (OU)
process~\citep{uhlenbeck1930} on the species tree, with analytic
covariance $\sigma^{2}/(2\alpha)\exp(-\alpha\, d_{\mathrm{tax}})$. The
Linnaean rank distance $d_{\mathrm{tax}}$ between species
(\cref{fig:teaser}), derived
from standard avian taxonomy~\citep{ioc-bird-list}, is a biologically
grounded quantity, available globally from existing classification,
and informed at the cross-order level by the whole-genome avian
phylogeny~\citep{jarvis2014}, precisely the signal a
generalization-bound estimator should be using.

\paragraph{Our approach: a taxonomic OU prior on the AVES shape code.}
We propose a \textbf{taxonomic Ornstein--Uhlenbeck prior} on the
AVES~\citep{aves2021} shape code $\beta$ (the low-dimensional shape
coefficients of the AVES mesh) that replaces the
$\mathcal{L}_{\mathrm{con}}$ used by AniMer+~\citep{animerplus2025}.
Per mini-batch, we look up pairwise Linnaean rank
distances from a fixed Linnaean rank distance matrix~\citep{ioc-bird-list}, assemble the
analytic OU covariance $\Sigma_{ij} = (\sigma^{2}/2\alpha)\exp(-\alpha
d_{\mathrm{tax}}(i,j))$, and add a multivariate-Gaussian negative
log-likelihood (\cref{eq:lou}) to the AniMer+ objective. The selection
strength $\alpha$ and the noise scale $\sigma$ are learnable scalars,
parameterized through softplus~\citep{dugas2001incorporating}, and are
the only new covariance parameters introduced. Conceptually, the OU prior is a
\emph{generative continuous extension} of the metric-learning
$\mathcal{L}_{\mathrm{con}}$ used by AniMer+: in the limit
$\alpha\!\to\!\infty$ with a block taxonomy that collapses each family
to a point, $\Sigma$ becomes a family-indicator block matrix and
$\mathcal{L}_{\mathrm{OU}}$ degenerates into a Gaussian form
structurally analogous to the family contrastive objective. The
analogy is structural: SupCon $L_2$-normalizes its feature space and
the Gaussian negative log-likelihood (NLL) does not. Finite $\alpha$ (our learned regime)
instead allows graded cross-family leakage and continuous within-family
granularity, exchanging a binary same-family indicator for a
rank-structured Gaussian likelihood with $\binom{17}{2}\!=\!136$
pairwise values. We summarize our contributions as follows:

\begin{itemize}
\item A taxonomic Ornstein--Uhlenbeck~\citep{uhlenbeck1930} prior on
  the AVES~\citep{aves2021} shape code that generalizes the family
  contrastive signal of AniMer+~\citep{animerplus2025} into a
  continuous, generative form indexed by Linnaean rank distance.
\item The proposed method improves over AniMer+ on both metrics of
  the unseen-species CowBird benchmark and on CtrlAVES3D, while
  remaining comparable on seen-species CUB.
\item A controlled permutation ablation isolates taxonomic structure
  from generic Gaussian smoothing.
\item A rank-granularity ablation identifies cross-order Linnaean
  distinctions, the structure beyond a family-vs-not indicator, as
  the dominant source of the unseen-species gain.
\end{itemize}

\section{Related Work}
\label{sec:related}

\paragraph{Animal 3D pose and shape.} Parametric mesh models---SMAL~\citep{smal2017} (a multi-animal extension of SMPL~\citep{smpl2015}) for quadrupeds and AVES~\citep{aves2021} for birds---are the dominant approach for animal 3D reconstruction. Category-specific mesh reconstruction~\citep{cmr2018} and wild-animal fitting~\citep{smalr2018,threed_safari2019} established the learning-based paradigm. Single-image regressors have since evolved from early
bird-specific reconstruction~\citep{badger2020} through breed-aware
(BARC~\citep{barc2022}) and mixture-of-Gaussians (WLDO~\citep{wldo2020})
shape priors to category-level (MagicPony~\citep{magicpony2023}) and
pan-category (3D-Fauna~\citep{3dfauna2024}) models. For birds,
AniMer+~\citep{animerplus2025} (extending the mammal-only
AniMer~\citep{animer2025}) is the current state of the art: a
family-aware mixture-of-experts~\citep{riquelme2021scalingmoes}
transformer~\citep{dosovitskiy2021an} trained jointly on mammals and
birds with a family-level supervised-contrastive~\citep{supcon2020}
$\mathcal{L}_{\mathrm{con}}$ term. We keep the
AniMer+ backbone but replace the discrete
$\mathcal{L}_{\mathrm{con}}$ used by AniMer+ with a continuous rank-structured
generative prior, and show that continuous taxonomic structure
matters, both theoretically and empirically.
$\mathcal{L}_{\mathrm{con}}$ is the structural analogue of our prior in
the degenerate $\alpha\!\to\!\infty$ case (\cref{sec:method:limit}),
and a controlled permutation ablation tests the taxonomic content.

\paragraph{Taxonomic regularization and phylogenetic priors in ML.}
Hierarchical classification~\citep{bertinetto2020,barz2019} embeds
class-level structure into learning objectives, and image-based methods
exploit taxonomy discretely, including the breed triplet loss of
BARC~\citep{barc2022} and the family-level supervised
contrastive~\citep{supcon2020} loss of AniMer+; the latter two are
pair losses with a 1-bit-per-pair information budget. Such losses can
be read as implicit class-conditional Gaussians with a block-diagonal covariance (one
block per family). The OU~\citep{uhlenbeck1930} process as a model
of continuous trait evolution on a tree is classical in phylogenetic
comparative methods~\citep{felsenstein1985,hansen1997,butler2004};
tree-structured OU priors have appeared in
Draupnir~\citep{draupnir2022}, a VAE for ancestral protein sequences,
but to our knowledge not in 3D articulated shape estimation. Our prior
replaces the block-diagonal with a dense, rank-structured $\Sigma$
indexed by a Linnaean~\citep{ioc-bird-list} rank distance,
retaining the discrete class-block as the $\alpha\!\to\!\infty$
degenerate case (\cref{sec:method:limit}).

\paragraph{Empirical phylogenetic priors on bird shape.}
\citet{aves2021} reported a strong phylogenetic signal
(Pagel's~\citep{pagel1999} $\lambda{=}0.99$, $p{<}10^{-4}$) on
low-dimensional embeddings of the AVES shape principal components, evidence that
articulated avian shape codes carry continuous phylogenetic structure
even under per-image keypoint supervision. The multi-modal Gaussian
shape prior of WLDO~\citep{wldo2020} is the discrete-mixture analogue
of our continuous OU~\citep{uhlenbeck1930} kernel: a mixture of $M$ Gaussians
($M{=}10$ in WLDO) whose block structure is recovered as the
$\alpha\!\to\!\infty$ block-taxonomy limit of our
prior (\cref{sec:method:limit}). Our prior generalizes both lines
along two axes simultaneously, moving from a discrete taxonomy to a
continuous one and from a metric-based formulation to a generative
likelihood.

\section{Method}
\label{sec:method}

\begin{figure}[t]
\centering
\includegraphics[width=1.0\linewidth]{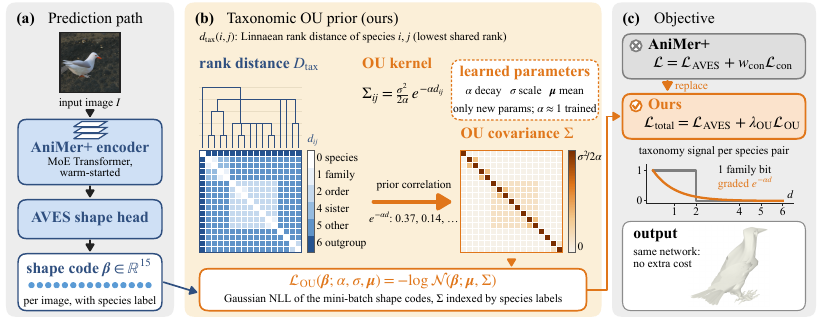}
\vspace{-0.6cm}
\caption{\textbf{Method at a glance.}
\textbf{(a)} The AniMer+~\citep{animerplus2025} encoder produces
AVES~\citep{aves2021} shape codes
$\bm{\beta}$ from a single image.
\textbf{(b)} Across species, $\bm{\beta}$ is regularized by
an OU~\citep{uhlenbeck1930} prior parameterized by learnable
$(\alpha, \sigma, \bm{\mu})$ over a
fixed Linnaean~\citep{ioc-bird-list} rank distance matrix; the NLL of this
Gaussian is $\mathcal{L}_{\mathrm{OU}}$.
\textbf{(c)} $\mathcal{L}_{\mathrm{OU}}$ replaces the family-level
contrastive term of AniMer+ ($w_{\mathrm{con}}{=}0$), so that on birds
$\mathcal{L}_{\mathrm{total}} = \mathcal{L}_{\mathrm{AVES}}
+ \lambda_{\mathrm{OU}} \mathcal{L}_{\mathrm{OU}}$.
Blue: the existing network and fixed taxonomy. Orange: what we add.}
\label{fig:architecture}
\end{figure}

\paragraph{Overview.}
\Cref{fig:architecture} sketches the method.
\textbf{(a)} The AniMer+~\citep{animerplus2025} encoder,
initialized from the released checkpoint, produces per-image
shape codes $\bm{\beta}$ for AVES~\citep{aves2021}, a multi-species
parametric bird mesh whose shape is controlled by a $15$-dimensional
coefficient vector.
\textbf{(b)} Across the species in our training split, we model the
joint distribution of $\bm{\beta}$ as a multivariate Gaussian whose
covariance follows an Ornstein--Uhlenbeck (OU)
kernel~\citep{uhlenbeck1930,hansen1997,butler2004} over a fixed
Linnaean~\citep{ioc-bird-list} rank distance matrix
$D_{\mathrm{tax}}$. The kernel is parameterized by two learnable
scalars $(\alpha,\sigma)$ and a learnable species-mean $\mu$,
yielding the OU regularizer $\mathcal{L}_{\mathrm{OU}}$.
\textbf{(c)} $\mathcal{L}_{\mathrm{OU}}$ replaces AniMer+'s family contrastive
term and is added to the original task losses at strength
$\lambda_{\mathrm{OU}}$. Restoring the contrastive term and setting
$\lambda_{\mathrm{OU}}\!=\!0$ recovers AniMer+ to floating-point equality,
so the OU contribution is isolated in a single hyperparameter.

The remainder of this section reviews the OU process and the
taxonomic distance it consumes (\cref{sec:method:prelim}), formalizes
the OU kernel on shape codes (\cref{sec:method:ou}) and the resulting
NLL loss (\cref{sec:method:nll}), discusses the relationship to family
supervised contrastive~\citep{supcon2020} (\cref{sec:method:limit}),
and specifies how the prior integrates with AniMer+
(\cref{sec:method:integration}).

\subsection{Preliminaries: OU process and taxonomic distance}
\label{sec:method:prelim}

\paragraph{Intuition.}
Closely related species tend to have similar body shapes, and the
similarity fades as the relationship becomes more distant. An OU
process is a standard model in comparative phylogenetics for a
continuous trait that evolves along a species tree while being
pulled toward an optimum; under it the trait values of two species
are correlated, and the correlation decays exponentially with their
distance on the tree. The Linnaean rank distance supplies that
distance from standard taxonomy alone, as a small integer that grows
as the pair's shared group moves up from species to family, order,
and beyond. The resulting covariance matrix over species is the
Gaussian prior we place on the AVES shape code.

\paragraph{OU process.}
The Ornstein--Uhlenbeck process~\citep{uhlenbeck1930} is a stationary
mean-reverting stochastic process whose covariance between two time
points decays exponentially with their separation. In comparative
phylogenetics, OU processes have long been the standard model of
continuous trait evolution on a species
tree~\citep{hansen1997,butler2004,blomberg2003}: the parameter
$\alpha$ controls selection strength (pull toward an optimum) and
$\sigma$ scales the diffusion noise. Setting $\alpha\!\to\!0$ recovers the
Brownian-motion limit with covariance proportional to shared branch
length~\citep{felsenstein1985}; $\alpha > 0$ bounds the stationary
variance to $\sigma^{2}/(2\alpha)$.

\paragraph{Taxonomic distance.}
An OU model takes as input a pairwise relatedness distance
$d_{\mathrm{tax}}(i,j)$ between species $i$ and $j$. We use the
\emph{Linnaean rank distance} based on standard avian
taxonomy~\citep{ioc-bird-list}: $d_{\mathrm{tax}}(i,j) = 0$ for the
same species, $1$ for the same family, $2$ for the same order, and
$4$--$6$ for cross-order pairs. The cross-order values are informed by
the whole-genome avian phylogeny~\citep{jarvis2014}, and $5$--$6$ are
ordinal design choices (\cref{sec:sup:dtax_matrix}).
Restricted to the 17 species in CtrlAVES3D~\citep{ctrlaves3d2024} this
yields a fixed, non-learnable $17{\times}17$ float32 matrix
$D_{\mathrm{tax}}$ fed into the OU kernel below.

\paragraph{Why Linnaean ranks rather than a molecular phylogeny?}
The Jetz et al.\ avian supertree~\citep{jetz2012} would seem the
natural input to an OU kernel since comparative phylogenetics
classically consumes patristic distances (path lengths through the
tree). Two scale-conditioning arguments justify the discrete rank distance instead.
\textbf{(i) Conditioning.} On this $17$-species subset Jetz patristic
distances span the range $[18.6, 201.2]$ million years;
 for $\exp(-\alpha\, d)$ to express
meaningful cross-family correlation on that scale, $\alpha$ must live
near $10^{-2}$, leaving a near-diagonal covariance in the regime where
our optimizer converges ($\alpha\!\sim\!1$). Linnaean~\citep{ioc-bird-list}
ranks ($\{1,2,4,5,6\}$ off-diagonal) lie on the same order as the
converged $\alpha$, yielding meaningful across-species sharing without rescaling.
\textbf{(ii) Signal preservation.} Linnaean ranks correlate
strongly with Jetz patristic distances on this subset (Pearson $0.85$,
Spearman $0.76$), so the discrete rank kernel keeps most of the
biological signal of a continuous molecular kernel while gaining a
much better-conditioned covariance. The Jetz tree is used only for
this check and for the evaluation in \cref{sec:phylo_calibration};
training sees $D_{\mathrm{tax}}$ alone. Further discussion of the
conditioning argument and of length-scale-normalized variants is
in~\cref{sec:sup:dtax:why_discrete,sec:sup:future_work}.

\subsection{OU kernel on shape codes}
\label{sec:method:ou}

\paragraph{Why this kernel.}
We want a species-level prior that (a) places shape codes of related
species near each other in $\mathbb{R}^{15}$, (b) does so with a
strength that decays smoothly with taxonomic distance rather than at
a hard family cut, and (c) keeps a global ``how much can shape vary''
knob separate from the relational ``how similar are sisters'' knob.
The OU kernel satisfies all three by construction.

Instantiating the OU process on the species taxonomy gives a stationary
multivariate Gaussian over the tips whose $(i,j)$ covariance
is~\citep{butler2004,hansen1997}:
\begin{equation}
\mathrm{Cov}(\beta_i, \beta_j) \;=\; \frac{\sigma^2}{2\alpha}
  \exp\!\big(-\alpha\, d_{\mathrm{tax}}(i,j)\big),
\label{eq:ou_cov}
\end{equation}
where $d_{\mathrm{tax}}(i,j)$ is the Linnaean rank distance from
\cref{sec:method:prelim}. The prefactor $\sigma^{2}/(2\alpha)$ is
the stationary variance of the OU process and sets the prior's
marginal scale on each species; the exponential factor $\exp(-\alpha
d)$ encodes the relational decay with taxonomic distance. Large
$\alpha$ collapses the kernel to sharp family-like blocks; small
$\alpha$ smears similarity across the entire taxonomy. Both scalars
are learned via softplus parameterizations~\citep{dugas2001incorporating}
enforcing positivity (with floor $\alpha_{\min}=10^{-3}$) and
initialized at $\alpha_0=\sigma_0=1$ (the raw pre-softplus parameter
is set to the inverse softplus of $1$).

\subsection{NLL loss}
\label{sec:method:nll}

\paragraph{Why an NLL.}
Treating $\mathcal{L}_{\mathrm{OU}}$ as the negative log-likelihood of
a Gaussian (rather than, say, a metric-learning surrogate) gives the
prior a generative interpretation that downstream analysis exploits:
the same $\Sigma$ that appears in the loss is the Gaussian-process
(GP) covariance whose maximum-a-posteriori (MAP) reading we use in
\cref{sec:sup:theory:map}. Generative form also makes the
$\alpha\!\to\!\infty$ limit comparison to discrete family contrastive
clean (\cref{sec:method:limit}).

Let $\bm{\beta} \in \mathbb{R}^{B \times D}$ be the AVES shape codes
predicted for a mini-batch of $B$ instances (here $D = 15$, taxonomy
size $N = 17$), and let $\Sigma_{ij} =
(\sigma^2/2\alpha)\exp(-\alpha\,d_{\mathrm{tax}}(s_i, s_j)) +
\varepsilon\mathbf{1}[i=j]$ with jitter $\varepsilon = 10^{-6}$. Treating
the $D$ shape dimensions as independent samples from a shared species
covariance, the prior, written as a loss, is
\begin{equation}
\mathcal{L}_{\mathrm{OU}} \;=\; \frac{D}{2}\log|\Sigma| \;+\;
   \frac{1}{2}\sum_{d=1}^{D}(\bm{\beta}_{\cdot, d} - \mu)^{\top}\Sigma^{-1}
   (\bm{\beta}_{\cdot, d} - \mu)
\label{eq:lou}
\end{equation}
The mean $\mu \in \mathbb{R}^D$ is also learnable, so the prior pulls
$\bm{\beta}$ toward the empirical species mean rather than toward a
fixed origin. Numerical implementation
is detailed in
\cref{sec:sup:method_extended}.

Before turning to the integration with AniMer+, we first relate this
prior to the family supervised-contrastive loss it
generalizes---examining the kernel under the degenerate
$\alpha\!\to\!\infty$ limit makes the connection precise.

\subsection{Relationship to family supervised contrastive}
\label{sec:method:limit}

\paragraph{Intuition: what does $\alpha\!\to\!\infty$ mean?}
Recall that $\alpha$ controls how fast the kernel $\exp(-\alpha\,d)$
decays with taxonomic distance. Sending $\alpha$ to infinity is the
limit in which the prior treats every same-family pair as
\emph{interchangeable} ($d{=}0$ kept, all $d>0$ off-diagonals driven
to zero) and every cross-family pair as \emph{unrelated} (correlation
exactly zero). This is precisely the binary view of taxonomy that
the family-level supervised-contrastive loss $\mathcal{L}_{\mathrm{con}}$
of AniMer+~\citep{animerplus2025} encodes: same-family pairs pulled
together, all other pairs pushed apart, with no graded gradation.
We make this connection precise below.

\paragraph{Limit form of the kernel.}
Under the block-taxonomy convention $d_{\mathrm{tax}}(i,j) = 0$ for
same-family pairs and $d_{\mathrm{tax}}(i,j) > 0$ otherwise, taking
$\alpha\!\to\!\infty$ collapses the OU kernel
$(\sigma^2/2\alpha)\exp(-\alpha\,d_{\mathrm{tax}}(i,j))$ to a
family-indicator block matrix scaled by the stationary variance
$\sigma^2/(2\alpha)$, held fixed in the limit. The block-indicator $\Sigma$ pulls
same-family $\bm{\beta}$ toward a shared family mean while
leaving different-family codes uncorrelated, the same qualitative
effect that supervised contrastive learning~\citep{supcon2020} induces in
feature space. Within a family, the joint marginal of the codes is a
degenerate Gaussian along the family direction. In this sense $\mathcal{L}_{\mathrm{OU}}$ supplies a
Gaussian probabilistic relaxation of the binary same-family
indicator that $\mathcal{L}_{\mathrm{con}}$ encodes.

\paragraph{Caveats: the analogy is structural, not literal.}
Two technical differences prevent equality in the limit.
\textbf{(i) Singular covariance.} At $\alpha\!\to\!\infty$ the block-indicator
$\Sigma$ is rank-deficient, so the strict Gaussian NLL is ill-defined;
one must take a pseudo-inverse with jitter $\varepsilon\!>\!0$
(we use $\varepsilon=10^{-6}$) and analyze the limit as
$\varepsilon\!\to\!0$. \textbf{(ii) Feature space.}
SupCon~\citep{supcon2020} operates on $L_2$-normalized class-token
features through a cosine-softmax objective, whereas
$\mathcal{L}_{\mathrm{OU}}$ operates on raw shape
codes $\bm{\beta}\in\mathbb{R}^{15}$ through a quadratic Mahalanobis
form. The two losses are best read as \emph{structurally analogous
regularizers acting on different representations}, both encouraging
same-family proximity. A proof sketch of the limit and the
structural comparison are in
\cref{sec:sup:theory_refinement}.

\paragraph{Why this matters.}
Our optimizer converges to finite $\alpha\!\sim\!1$, not infinity. At
that operating point, within-family pairs have correlation
$\exp(-\alpha)\!\approx\!0.37$, same-order pairs
$\exp(-2\alpha)\!\approx\!0.14$, and so on, a graded similarity
decay across the taxonomic hierarchy. The $\alpha\!\to\!\infty$
analogy clarifies what SupCon~\citep{supcon2020} is doing in the
limit; the finite-$\alpha$ regime our prior actually inhabits is
what our paper studies, and what $\mathcal{L}_{\mathrm{con}}$ cannot
represent.

\subsection{Integration with AniMer+}
\label{sec:method:integration}

\paragraph{Existing AniMer+ pipeline.}
The training loop of AniMer+~\citep{animerplus2025} returns a base
objective $\mathcal{L}_{\mathrm{base}} = \mathcal{L}_{\mathrm{AVES}}
+ \mathcal{L}_{\mathrm{SMAL}} + w_{\mathrm{con}}\,
\mathcal{L}_{\mathrm{con}}$ that combines per-keypoint, silhouette,
and shape-and-pose regression terms with the family
supervised-contrastive regularizer~\citep{supcon2020}.
$\mathcal{L}_{\mathrm{AVES}}$ and $\mathcal{L}_{\mathrm{SMAL}}$ are
the avian and mammalian
branches~\citep{aves2021,smal2017,smpl2015,kanazawa2018hmr}, and
$\mathcal{L}_{\mathrm{SMAL}}$ is inactive in our avian-only
configuration. The encoder is a mixture-of-experts
(MoE) Vision Transformer
(ViT)~\citep{dosovitskiy2021an,riquelme2021scalingmoes} pretrained
across mammal and avian datasets.

\paragraph{Our addition.}
We augment $\mathcal{L}_{\mathrm{base}}$ with our OU prior at strength
$\lambda_{\mathrm{OU}}$:
\begin{equation}
\mathcal{L}_{\mathrm{total}} \;=\; \mathcal{L}_{\mathrm{base}} \;+\;
  \lambda_{\mathrm{OU}}\,\mathcal{L}_{\mathrm{OU}}(\bm{\beta}_{\mathrm{pred}},\,
  f_{\mathrm{batch}},\, d_{\mathrm{tax}}),
\end{equation}
where $\bm{\beta}_{\mathrm{pred}}$ is the AVES~\citep{aves2021}-branch
shape code and $f_{\mathrm{batch}}$ is the per-sample species index.
In our best setting we set $w_{\mathrm{con}} \leftarrow 0$, so the OU
prior is the sole taxonomy-aware regularizer. With
$w_{\mathrm{con}}$ left at its original value, setting
$\lambda_{\mathrm{OU}} = 0$ recovers the
AniMer+~\citep{animerplus2025} training loop to floating-point equality,
isolating the OU contribution to a single hyperparameter.

\paragraph{Why the OU group needs a separate learning rate.}
We place the OU scalars $(\alpha,\sigma,\mu)$ in their own AdamW
parameter group whose learning rate (LR) is $k$ times the backbone
LR; we call $k$ the OU-LR multiplier.
Eval is nearly $k$-invariant at a given $\lambda_{\mathrm{OU}}$
across $k\in\{1,10,30\}$,
 confirming that the
OU prior's eval contribution comes primarily from the regularization
\emph{shape} at the initialization $(\alpha_0,\sigma_0)=(1,1)$
rather than from learned scalar updates, independent evidence against an
$(\alpha,\sigma)$-fitting artifact. We use $k\!=\!100$ in all
headline runs.

\begin{table}[t]
\centering
\small
\caption{\textbf{Comparison against AniMer+~\citep{animerplus2025} on the
avian benchmarks.} CUB~\citep{cub2011} and the
CtrlAVES3D~\citep{ctrlaves3d2024} test split evaluate \emph{seen}-species;
CowBird~\citep{badger2020} is the \emph{unseen}-species
out-of-distribution (OOD) benchmark.
For ours we report the mean $\pm$ sample standard deviation over
five random seeds.
AniMer-A~\citep{animer2025} is the avian-only variant; AniMer+ and
ours are unified bird+mammal models; AVES and HMR2.0 are the baselines
reported by AniMer+.
}
\label{tab:main}
\scalebox{0.95}{
\begin{tabular}{lccccc}
\toprule
 & \multicolumn{2}{c}{CUB} & \multicolumn{2}{c}{CowBird} & CtrlAVES3D \\
\cmidrule(lr){2-3}\cmidrule(lr){4-5}\cmidrule(lr){6-6}
Method & AUC $\uparrow$ & PCK $\uparrow$ & AUC $\uparrow$ & PCK $\uparrow$ & AUC $\uparrow$ \\
\midrule
AVES~\citep{aves2021}          & 79.70 & 25.20 & 46.10 &  5.60 & 85.90 \\
HMR2.0~\citep{hmr2}            & 87.70 & 50.00 & 59.10 & 12.20 & 91.60 \\
\midrule
AniMer-A~\citep{animer2025}    & 91.90 & 74.80 & 65.70 & 19.40 & 92.90 \\
\midrule
AniMer+~\citep{animerplus2025} & $\mathbf{89.50}$ & $\mathbf{59.40}$ & 59.30 & 16.40 & 93.00 \\
\textbf{Ours} & $89.30 \pm 0.02$ & $57.84 \pm 0.09$ & $\mathbf{60.80 \pm 0.10}$ & $\mathbf{18.86 \pm 0.27}$ & $\mathbf{93.05 \pm 0.01}$ \\
\bottomrule
\end{tabular}
}
\end{table}

\section{Experiments}
\label{sec:experiments}

\Cref{sec:method:train} fixes the training and evaluation setup.
\Cref{sec:main_table} reports the headline result.
\Cref{sec:permutation_ablation} probes whether the gain is genuinely taxonomic
through a label-permuted control, and \cref{sec:family_block}
sharpens the question into a rank-granularity ablation.
\Cref{sec:vertex} reports 3D vertex-level errors as a complementary
check, and \cref{sec:phylo_calibration} verifies that the OU prior
tightens the shape codes in $\bm{\beta}$-space without distorting
their phylogenetic signal.
Qualitative comparisons follow in \cref{sec:qualitative_main}.

\subsection{Training and evaluation setup}
\label{sec:method:train}

We fine-tune from the released AniMer+~\citep{animerplus2025}
checkpoint on the $6\,464$-image train split of
CtrlAVES3D~\citep{ctrlaves3d2024} for $30{,}000$ steps with
AdamW~\citep{loshchilov2017decoupled} and a cosine learning-rate
schedule~\citep{loshchilov2017sgdr}. The training data covers $17$
avian species; evaluation uses three benchmarks following the protocol of
AniMer+ (Table~VI): CUB-200-2011~\citep{cub2011} (CUB) for
seen-species fine localization, CowBird~\citep{badger2020} for
unseen-species OOD, and the CtrlAVES3D test split for 3D
reconstruction. Following AniMer+, we report the percentage of correct
keypoints (PCK) at threshold $0.1$ (PCK@0.1), i.e.\ the fraction of
predicted 2D keypoints whose normalized distance to the ground truth
is below $0.1$, and the area under the PCK curve (AUC) as the
threshold sweeps from $0$ to $1$. Our model is warm-started from
the unified AniMer+~\citep{animerplus2025} bird+mammal checkpoint and fine-tuned on the
avian split (mammal-branch weight set to zero). Headline numbers
(\cref{tab:main,tab:vertex}) use $\lambda_{\mathrm{OU}}=10^{-5}$ with
the cosine schedule; the real-vs-permuted and rank-granularity
ablations (\cref{fig:permutation_null,tab:family_block}) use a
stronger setting, $\lambda_{\mathrm{OU}}=10^{-4}$ with a constant
learning rate, so that the taxonomic content of $D_{\mathrm{tax}}$ is
expressed strongly enough for the contrast to be visible. Full implementation details (fp32
training, two-group AdamW with OU-LR multiplier, gradient
clipping, jitter and softplus~\citep{dugas2001incorporating}
initialization, the permutation-matrix control protocol, and the
integration with AniMer+'s training loop) are in \cref{sec:sup:training_impl}.

\subsection{Main result: unseen-species gains over AniMer+}
\label{sec:main_table}

\Cref{tab:main} compares the proposed method against the avian
numbers reported by AniMer+~\citep{animerplus2025}. Our method
improves over the AniMer+ paper on both unseen-species CowBird
metrics ($+1.5$ AUC, $+2.5$ PCK@0.1) and on CtrlAVES3D, and remains
comparable on seen-species CUB.
Improvements concentrate on the unseen-species
CowBird~\citep{badger2020} benchmark, where the cross-species
borrowing supplied by the OU prior is most useful.
AniMer-A~\citep{animer2025} is a bird-only specialist that sacrifices
the unified multi-domain coverage of AniMer+; the like-for-like
comparison is therefore against the unified AniMer+ baseline.

\subsection{Real vs.\ permuted taxonomy: is the gain genuinely taxonomic?}
\label{sec:permutation_ablation}

\begin{SCfigure}[1.0][t]
\centering
\includegraphics[width=0.6\linewidth]{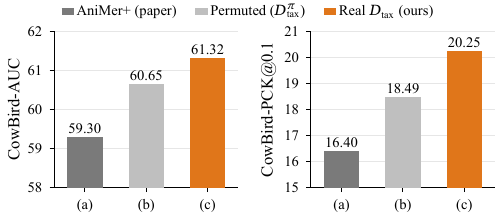}
\caption{\textbf{Real vs.\ permuted taxonomy ablation.}
CowBird~\citep{badger2020} AUC and PCK@0.1 under three
configurations: (a) the AniMer+~\citep{animerplus2025} baseline,
(b) the OU prior with a label-permuted distance matrix, and (c) the
OU prior with the real Linnaean~\citep{ioc-bird-list} distance
matrix (both OU runs at $\lambda_{\mathrm{OU}}=10^{-4}$, constant
learning rate; \cref{sec:method:train}).}
\label{fig:permutation_null}
\end{SCfigure}

A regularizer that improves OOD performance by injecting any
shape-code structure, even random structure, would not constitute
evidence for the taxonomic hypothesis. \Cref{fig:permutation_null}
addresses this concern by keeping the training configuration fixed and changing only
whether the OU kernel sees the biologically-correct distance matrix
or a label-permuted control. The permutation preserves the OU-kernel
spectrum exactly, with the same covariance eigenvalues,
but destroys the correspondence between species labels and their
taxonomy positions.

The real-vs-permuted comparison is the key result. The OU prior with the
real Linnaean~\citep{ioc-bird-list} distance matrix outperforms the
permuted control on both CowBird-AUC and CowBird-PCK@0.1~\citep{badger2020}.
The AniMer+~\citep{animerplus2025} bar is the value reported in the
original paper rather than a run from our pipeline, so its gap to the
OU runs is not a controlled comparison and is shown for context.
The OU prior is therefore using \emph{which species are which}, not
merely the fact that an OU kernel was applied.
The same real-vs-permuted contrast holds on the $183$
CUB species outside the fine-tuning taxonomy, with species-clustered
confidence intervals excluding zero (\cref{sec:sup:unseen_cub}).

\subsection{Beyond family labels: rank-granularity ablation}
\label{sec:family_block}

\Cref{tab:family_block} sharpens the previous question into a
quantitative one: how much of the unseen-species gain depends on
information \emph{beyond} a family-vs-not indicator? This also
empirically tests the structural claim of \cref{sec:method:limit}:
the family-only $D_{\mathrm{tax}}$ row is the discrete
$\alpha\!\to\!\infty$ limit of our prior (analogous to family
SupCon), and the full Linnaean row exposes the graded cross-rank
structure that our finite-$\alpha$ regime was designed to capture.
We compare three progressively richer versions of the distance
matrix, namely a family-only block, an order-only refinement, and
the full Linnaean ranks, all under the same training configuration
as \cref{sec:permutation_ablation}.
The three rank-aware kernels separate cleanly along a single axis:
how much taxonomic detail beyond family membership is exposed to the
prior. Two readings stand out.

\begin{table}[t]
\centering
\small
\caption{\textbf{Rank-granularity ablation}: progressively richer distance
matrices on the same operating point as \cref{fig:permutation_null}. The
family-only kernel uses values in $\{0,1,5\}$ (same family vs.\
not), the order-only refinement uses $\{0,1,2,5\}$ (adds the
same-order distinction), and the full Linnaean kernel uses
$\{0,1,2,4,5,6\}$ (cross-order ordinals as the default).
All three kernels use $0$ for the same species and $1$ for the same
family, and differ only above the family level; genus-level structure
cannot be varied on this training set, because the $17$ species
contain no congeneric pair.
We use $5$ rather than $6$ as the cross-order placeholder for the
family-only and order-only kernels so that they share a single
cross-order bin, isolating the rank-granularity axis from changes
in cross-order distance scale.}
\scalebox{0.95}{
\label{tab:family_block}
\begin{tabular}{lccccc}
\toprule
 & \multicolumn{2}{c}{CUB~\citep{cub2011}} & \multicolumn{2}{c}{CowBird~\citep{badger2020}} & CtrlAVES3D~\citep{ctrlaves3d2024} \\
\cmidrule(lr){2-3}\cmidrule(lr){4-5}\cmidrule(lr){6-6}
$D_{\mathrm{tax}}$ variant & AUC $\uparrow$ & PCK $\uparrow$ & AUC $\uparrow$ & PCK $\uparrow$ & AUC $\uparrow$ \\
\midrule
permuted (null)                & 88.87 & 54.90 & 60.65 & 18.49 & 92.95 \\
family-only                    & 89.09 & 56.35 & 60.84 & 19.09 & 93.06 \\
order-only                     & 89.10 & 56.42 & 60.81 & 19.20 & 93.07 \\
\textbf{Linnaean (full, ours)} & 88.92 & 55.16 & $\mathbf{61.32}$ & $\mathbf{20.25}$ & 92.99 \\
\bottomrule
\end{tabular}
}
\end{table}

\paragraph{Cross-order ranks carry the unseen-species gain.}
The family-only and order-only kernels each recover only a fraction
of the CowBird~\citep{badger2020} improvement that the full
Linnaean~\citep{ioc-bird-list} kernel achieves;
the additional cross-order rank distinctions (e.g.\ separating the
duck order Anseriformes from the woodpecker order Piciformes) supply the dominant share of the
unseen-species gain. Concretely, ranks above the family level account
for about $66$\% of the CowBird-PCK improvement over the permuted
null, and cross-order ranks alone for about $60$\%. This is
exactly the structural information that a family-vs-not indicator
\emph{cannot} represent, and the empirical loading on these ranks is
why the title is ``beyond family labels''.

\paragraph{Seen vs.\ unseen trade is consistent with cross-species
borrowing.}
The CUB~\citep{cub2011}-PCK pattern is the converse: family-only and
order-only matrices retain more seen-species fine localization than
the full Linnaean~\citep{ioc-bird-list} kernel, which sacrifices a
small amount of CUB-PCK to amplify the
CowBird~\citep{badger2020} gain. This is the prior's intended trade-off:
finer cross-order granularity tightens cross-species mixing, which
helps unseen species at a small cost on seen-species fine
localization. The mechanism is a property of the \emph{structure}
of $D_{\mathrm{tax}}$, not a generic regularization effect.

\subsection{3D vertex-level errors on CtrlAVES3D}
\label{sec:vertex}

The headline table reports 2D AUC and PCK; for completeness we
additionally evaluate Procrustes-aligned~\citep{kanazawa2018hmr} 3D
errors on the CtrlAVES3D~\citep{ctrlaves3d2024} test
split.
\Cref{tab:vertex} reports the Procrustes-aligned mean per-joint and
per-vertex position errors (pa-MPJPE and pa-MPVPE) against the
AniMer+ released checkpoint on the same eval pipeline. The
2D-keypoint gain in \cref{tab:main} is reflected in the 3D mesh: the
OU prior reduces both pa-MPJPE and pa-MPVPE relative to the baseline
($\Delta\,\text{pa-MPJPE}{=}{-}5.72$,
$\Delta\,\text{pa-MPVPE}{=}{-}9.96$ in canonical units;
\cref{tab:vertex}).
The real-vs-permuted $\dphylo$ contrast shows the same sign as the
CowBird-PCK ablation in \cref{fig:permutation_null}
($\Delta\,\text{pa-MPVPE}{=}{-}1.33$ in favor of real
$\dphylo$).

\begin{table}[t]
\centering
\small
\caption{\textbf{Procrustes-aligned 3D errors on the CtrlAVES3D test split
($309$ images, $17$ species)}. pa-MPJPE denotes the mean per-joint position error;
pa-MPVPE denotes mean per-vertex position error. Lower is better. Ours is a single run (seed~$0$). Values
follow our canonical-frame scaling; see \cref{sec:sup:vertex_errors} for details.}
\label{tab:vertex}
\begin{tabular}{lcc}
\toprule
Variant & pa-MPJPE $\downarrow$ & pa-MPVPE $\downarrow$ \\
\midrule
AniMer+~\citep{animerplus2025}  & $655.61$ & $708.97$ \\
\textbf{AniMer+ \textit{w/} OU (ours)}                       & $\mathbf{649.89}$ & $\mathbf{699.01}$ \\
\bottomrule
\end{tabular}
\end{table}

\subsection{\texorpdfstring{What the prior changes in
$\bm{\beta}$-space}{What the prior changes in beta-space}}
\label{sec:phylo_calibration}

\begin{figure}[t]
\centering
\includegraphics[width=\linewidth]{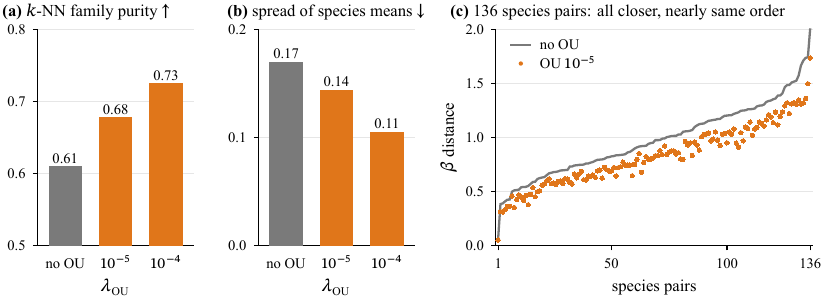}
\caption{\textbf{The prior tightens the shape codes without
distorting phylogenetic signal.} (a, b) What tightens: $k$-NN family
purity rises and the spread of the per-species mean $\bm{\beta}$ falls
with $\lambda_{\mathrm{OU}}$ on CtrlAVES3D~\citep{ctrlaves3d2024}
training codes; the $10^{-4}$ bars are a single constant-LR run.
(c) What is preserved: the $\bm{\beta}$ distance between the mean shape
codes of two species measures how different the model predicts their
shapes to be. All $136$ species pairs, sorted along the $x$-axis by this
distance without the prior (gray), move closer at
$\lambda_{\mathrm{OU}}{=}10^{-5}$ (orange, seed $0$) but stay in nearly
the same order.}
\label{fig:phylo_calibration}
\end{figure}

The shape codes of the pre-trained model already place related
species close together. Does the OU prior distort this structure? It
does not (\cref{fig:phylo_calibration}).

\textbf{What is preserved.} We compare the per-species mean codes
with the Jetz~\citep{jetz2012} phylogeny, an external tree that
training never sees; training uses only the Linnaean rank matrix of
\cref{sec:method}. Two standard statistics measure how well distance
on the tree predicts distance between codes: the Mantel
correlation~\citep{mantel1967} $r$ over the $136$ species pairs and
the multivariate Blomberg's $K$~\citep{blomberg2003,adams2014}. Both
are significantly above a $999$-permutation null for the released
AniMer+ checkpoint ($r{=}0.36$, $K_{\mathrm{mult}}{=}0.58$) and are
essentially unchanged with the prior ($r{=}0.352\pm0.002$,
$K_{\mathrm{mult}}{=}0.578\pm0.001$ over five seeds;
\cref{sec:sup:calibration}). The structure therefore comes from
pre-training, and the prior keeps it.

\textbf{What changes.} The codes become tighter. The fraction of each
code's ten nearest neighbors that share its family ($k$-NN purity)
rises from $0.61$ without the prior to $0.68$ at the headline strength
and $0.73$ at $\lambda_{\mathrm{OU}}{=}10^{-4}$, and the spread of the
per-species means shrinks ($0.17\to0.14\to0.11$). All $136$ species
pairs move closer, by $0.85\times$ on average, and their ordering is
nearly preserved (Spearman $0.99$), which is why $r$ and
$K_{\mathrm{mult}}$ barely move.
Real and permuted $D_{\mathrm{tax}}$ give the same $r$,
$K_{\mathrm{mult}}$ and spread at both strengths; they differ only in performance
(\cref{sec:permutation_ablation}).

In short, the prior contracts the shape space and tightens each
family's neighborhood, but does not rearrange the relatedness among
species that pre-training built.

\subsection{Qualitative comparison}
\label{sec:qualitative_main}

\begin{figure}[t]
\centering
\includegraphics[width=\linewidth]{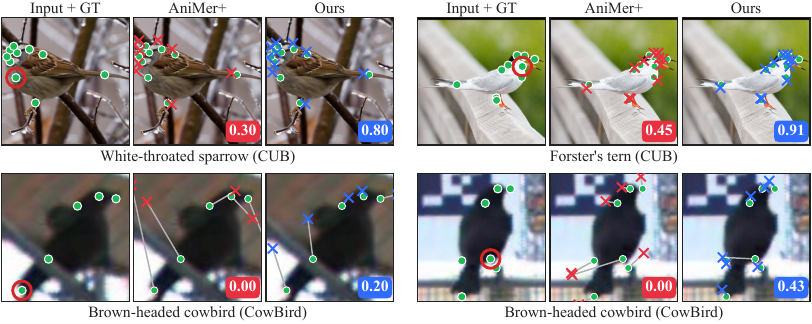}
\caption{\textbf{Qualitative comparison on two CUB~\citep{cub2011}
test images (top) and two CowBird~\citep{badger2020} test instances (bottom)}.
For each species, the first panel shows the input with the ground-truth
keypoints (green dots) and a red circle on the baseline's worst keypoint;
the retrained AniMer+ baseline (red crosses) and the OU fine-tune (blue
crosses) follow, and gray lines join each prediction to its ground-truth
keypoint. The badge is the per-image PCK@0.1.}
\label{fig:qualitative}
\end{figure}

\Cref{fig:qualitative} compares the AniMer+~\citep{animerplus2025}
baseline, retrained with its published recipe for the same $30{,}000$
steps, against the proposed OU fine-tune on two
CUB-200~\citep{cub2011} test images and two unseen-species
CowBird~\citep{badger2020} instances on which the baseline misplaces
several keypoints. In all four cases the OU fine-tune moves the
keypoint set toward the ground truth, raising per-image PCK@0.1 by
$0.45$ to $0.50$ on CUB and by $0.20$ to $0.43$ on CowBird. Further
paired examples are reported in
\cref{sec:sup:qualitative,sec:sup:cowbird_qualitative}.

\section{Conclusion}
\label{sec:conclusion}

We introduced a taxonomic Ornstein--Uhlenbeck~\citep{uhlenbeck1930}
prior on the AVES~\citep{aves2021} shape code for single-view avian
3D shape recovery. The prior generalizes the family supervised
contrastive~\citep{supcon2020} signal used by the prior state of the
art~\citep{animerplus2025} into a continuous, generative form indexed
by a Linnaean~\citep{ioc-bird-list} rank distance matrix, supplying
graded cross-species similarity that a binary same-family indicator
cannot represent. As a unit, the proposed
method improves over the prior state of the art on both metrics of the
unseen-species CowBird benchmark and on CtrlAVES3D while remaining
comparable on seen-species CUB, isolates the contribution of taxonomic
structure from generic Gaussian smoothing through a controlled
permutation ablation, and identifies cross-order rank distinctions,
the information that lies beyond a family-vs-not indicator, as the
dominant source of the unseen-species gain. Taken together, the
experiments support a biology-grounded inductive bias for long-tailed
avian shape recovery.

\paragraph{Limitations.}
The prior is evaluated on a $17$-species training taxonomy with
limited within-order density and no congeneric pair, so genus-level
structure is untested, and the unseen-species benchmark,
CowBird~\citep{badger2020}, covers a single species. The rank matrix
$D_{\mathrm{tax}}$ follows a single taxonomic
authority~\citep{ioc-bird-list}; we have not tested robustness to
competing resolutions of contested inter-order relationships, and
larger-scale use should regenerate the matrix as taxonomic consensus
evolves. The OU formulation is
not bird-specific, but extending it to other animal groups requires a
compatible parametric shape model, taxonomic annotations and
3D-labeled data, and how the gains behave at hundreds or thousands
of species remains to be established.

\begin{ack}
This work was supported by the National Research Foundation of Korea (NRF) grant funded by the Korea government (MSIT) (RS-2026-25598824);
by the NRF grant funded by the Korea government (MSIT) (No.\ RS-2024-00412141);
and by the International Science Business Belt R\&D Program funded by the Korea government (MSIT) (RS-2026-25548978).
The authors declare no competing interests.
\end{ack}

\clearpage
\small
\bibliographystyle{unsrtnat}
\bibliography{references}

\normalsize
\clearpage
\appendix

\renewcommand{\thesection}{S\arabic{section}}
\renewcommand{\thetable}{S\arabic{table}}
\renewcommand{\thefigure}{S\arabic{figure}}
\renewcommand{\theequation}{S\arabic{equation}}
\setcounter{section}{0}
\setcounter{table}{0}
\setcounter{figure}{0}
\setcounter{equation}{0}
\renewcommand{\theHfigure}{S\arabic{figure}}
\renewcommand{\theHtable}{S\arabic{table}}
\renewcommand{\theHequation}{S\arabic{equation}}
\crefalias{section}{appendix}
\crefalias{subsection}{subappendix}
\crefalias{subsubsection}{subsubappendix}

\thispagestyle{plain}
\vspace*{2em}

\begin{center}
{\large\bfseries Appendix of}\\[1.2em]
{\LARGE\bfseries Beyond Family Labels: A Taxonomic\\[0.25em]
Ornstein--Uhlenbeck Prior for Avian 3D Shape Recovery}\\[3em]
\rule{0.55\linewidth}{0.4pt}
\end{center}

\vspace{2.5em}
\noindent{\large\bfseries Contents}\\[1em]

\noindent\textbf{S1}\hspace{1.5em}\nameref{sec:sup:calibration}%
  \dotfill p.~\pageref{sec:sup:calibration}\\[0.45em]
\noindent\textbf{S2}\hspace{1.5em}\nameref{sec:sup:theory_refinement}%
  \dotfill p.~\pageref{sec:sup:theory_refinement}\\[0.45em]
\noindent\textbf{S3}\hspace{1.5em}\nameref{sec:sup:method_extended}%
  \dotfill p.~\pageref{sec:sup:method_extended}\\[0.45em]
\noindent\textbf{S4}\hspace{1.5em}\nameref{sec:sup:dtax_matrix}%
  \dotfill p.~\pageref{sec:sup:dtax_matrix}\\[0.45em]
\noindent\textbf{S5}\hspace{1.5em}\nameref{sec:sup:extended_experiments}%
  \dotfill p.~\pageref{sec:sup:extended_experiments}\\[0.45em]
\noindent\textbf{S6}\hspace{1.5em}\nameref{sec:sup:vertex_errors}%
  \dotfill p.~\pageref{sec:sup:vertex_errors}\\[0.45em]
\noindent\textbf{S7}\hspace{1.5em}\nameref{sec:sup:cowbird_qualitative}%
  \dotfill p.~\pageref{sec:sup:cowbird_qualitative}\\[0.45em]
\noindent\textbf{S8}\hspace{1.5em}\nameref{sec:sup:qualitative}%
  \dotfill p.~\pageref{sec:sup:qualitative}\\[0.45em]
\noindent\textbf{S9}\hspace{1.5em}\nameref{sec:sup:training_impl}%
  \dotfill p.~\pageref{sec:sup:training_impl}\\[0.45em]
\noindent\textbf{S10}\hspace{1.0em}\nameref{sec:sup:future_work}%
  \dotfill p.~\pageref{sec:sup:future_work}

\clearpage

\section{Calibration: phylogenetic signal in \texorpdfstring{$\bbeta$}{beta}-space}
\label{sec:sup:calibration}

\Cref{sec:phylo_calibration} reports that the OU prior leaves the
phylogenetic signal of the shape codes intact while tightening their
local structure. This appendix gives the underlying numbers
and two further cross-checks on the trained shape codes: under OU,
the codes keep their alignment with the Jetz tree, their family
$k$-NN purity rises, and their rank correlation with the discrete
Linnaean distance falls slightly.

\paragraph{Why these three diagnostics.} We ask what an
Ornstein--Uhlenbeck (OU)~\citep{uhlenbeck1930,hansen1997} prior
trained on a discrete Linnaean distance does to the geometry of the
codes.
The Mantel correlation~\citep{mantel1967} $r$ and the multivariate Blomberg's
$K$ ($K_{\mathrm{mult}}$)~\citep{blomberg2003,adams2014} measure how
strongly the per-species mean code $\bar{\bbeta}_s\in\mathbb{R}^D$
follows the Jetz~\citep{jetz2012} tree: $r$ correlates pairwise code
distance with patristic distance over the $136$ species pairs, and
$K_{\mathrm{mult}}$ compares the dispersion of the species means with
its Brownian-motion expectation under the tree covariance ($K{=}1$
at the expectation, $0$ for no phylogenetic signal). Spearman correlation
between the empirical $\bbeta$-distance matrix and a candidate
phylogenetic distance matrix probes the global rank ordering. The
candidates are the Linnaean rank $\dphylo$, its family- and
order-only collapses, the Jetz~\citep{jetz2012} patristic distance in
millions of years (My)\footnote{$1\,\mathrm{My}=10^{6}$ years.}, the same
distance length-scale-normalized, and label-permutation controls.
$k$-NN purity by Linnaean
family/order (\cref{tab:sup_cluster_purity}) probes local
neighborhood structure. The three diagnostics are sensitive to
different things, which is exactly why we report all of them.

\paragraph{Summary of evidence presented in this section.}

\begin{itemize}
\item \emph{Phylogenetic signal on the Jetz tree:} unchanged by the
prior ($K_{\mathrm{mult}}$ $0.581\to0.578\pm0.001$, Mantel $r$
$0.358\to0.352\pm0.002$ over five seeds; \cref{tab:sup_phylo_signal}),
and closely matched for real and permuted $D_{\mathrm{tax}}$.
\item \emph{Local cluster purity:} OU monotonically tightens local
within-family neighborhoods. Family $k$-NN purity rises from $0.610$
without the prior to $0.678$ at our headline configuration, and
further under stronger regularization.
\item \emph{Spearman global rank correlation:} drops marginally
($-0.012$ on Jetz, $-0.038$ on Linnaean) under OU.
\end{itemize}

Together, these results give a coherent picture: OU keeps the
continuous-tree alignment where pre-training put it (Mantel $r$,
$K_{\mathrm{mult}}$ and Jetz Spearman flat) while tightening local
within-family neighborhoods (purity up) and loosening the discrete
rank correspondence slightly (Linnaean Spearman down). This supports the paper's
$k$-invariance reading (\cref{sec:method:integration},
\cref{tab:sup_lrmult}) that OU's eval contribution comes from
optimization shaping rather than from a wholesale restructuring of
the latent shape manifold.

\paragraph{Setup.} For each trained variant, we run inference on the
6\,464-image CtrlAVES3D~\citep{ctrlaves3d2024} train split (which carries species labels for
all $N=17$ species) and aggregate the predicted shape codes into a
per-species mean $\bar{\bbeta}_s$, giving an $N{\times}D$ trait matrix
that we plug into the Mantel, $K_{\mathrm{mult}}$ and Spearman analyses. This
per-species mean is the right maximum-likelihood contraction under the OU model
(individuals within a species are exchangeable) but does discard
within-species variation that the model has no taxonomic signal
about.

\begin{table}[t]
\centering
\small
\caption{\textbf{Spearman correlation $\rho$ between $\bbeta$-distance and phylogenetic distance.}
Per-species means computed from inference on the 6464-image
CtrlAVES3D~\citep{ctrlaves3d2024} train split, $N{=}17$ species,
$D{=}15$ shape dims; correlated against various phylogenetic distance
matrices across six variants.
\textbf{Bold} = the strongest non-permuted phylogeny within each row.
The Jetz~\citep{jetz2012} patristic distance dominates the rank in every variant.}
\label{tab:sup_beta_phylo_correlation}
\resizebox{\linewidth}{!}{%
\begin{tabular}{lcccccccc}
\toprule
Variant & Linnaean~\citep{ioc-bird-list} & Family & Order & Jetz~\citep{jetz2012} (My) & Jetz norm & Perm ($s_{42}$) & Perm ($s_{101}$) & Perm ($s_{202}$) \\
\midrule
AniMer+~\citep{animerplus2025} retrain @30K (no OU)    & $+0.106$ & $+0.119$ & $+0.166$ & $\mathbf{+0.358}$ & $+0.358$ & $+0.138$ & $-0.152$ & $-0.201$ \\
Cosine LR (no OU)               & $+0.105$ & $+0.116$ & $+0.167$ & $\mathbf{+0.354}$ & $+0.354$ & $+0.136$ & $-0.154$ & $-0.202$ \\
Constant LR + OU $\lambda{=}10^{-5}$        & $+0.070$ & $+0.121$ & $+0.136$ & $\mathbf{+0.354}$ & $+0.354$ & $+0.129$ & $-0.164$ & $-0.189$ \\
Cosine LR + OU $\lambda{=}10^{-5}$ (\textbf{ours}) & $+0.068$ & $+0.119$ & $+0.135$ & $\mathbf{+0.346}$ & $+0.346$ & $+0.123$ & $-0.170$ & $-0.189$ \\
\midrule
\multicolumn{9}{l}{\emph{Stronger OU regime ($\lambda{=}10^{-4}$, single seed)}} \\
\midrule
$\lambda{=}10^{-4}$ + real $\dphylo$ & $+0.030$ & $+0.129$ & $+0.094$ & $\mathbf{+0.330}$ & $+0.330$ & $+0.080$ & $-0.174$ & $-0.154$ \\
$\lambda{=}10^{-4}$ + perm $\dphylo$ & $+0.032$ & $+0.131$ & $+0.106$ & $\mathbf{+0.339}$ & $+0.339$ & $+0.081$ & $-0.171$ & $-0.153$ \\
\bottomrule
\end{tabular}%
}
\end{table}

\paragraph{Three observations.}

\textit{(a) AniMer+~\citep{animerplus2025} baseline already carries
phylogenetic signal} ($\rho_{\text{Jetz}}{=}{+}0.358$). The signal is
already present in the released checkpoint
(\cref{tab:sup_phylo_signal}), so it comes from pre-training rather
than from the prior.

\textit{(b) Spearman rank correlation against the Jetz tree is
essentially unchanged} (margin: $-0.012$ for cosine $+$ OU vs.\
AniMer+ retrain). Spearman is invariant to monotone re-scalings of
pairwise distances, so this flat result is consistent with the
near-uniform contraction in \cref{fig:phylo_calibration}(c): the
species means move closer without re-ordering the relationships
already there in the baseline.
Family $k$-NN purity (\cref{tab:sup_cluster_purity}), a local
statistic, rises while the global statistics (Spearman, Mantel $r$,
$K_{\mathrm{mult}}$) stay flat. Flat global ordering and higher local
purity are therefore complementary, not contradictory.

\textit{(c) OU-trained codes are slightly less aligned with the
discrete Linnaean rank distance}
($\rho_{\text{Linnaean}}$: $+0.068$ for cosine + OU vs.\ $+0.106$
baseline). The permuted control shows the same drop ($+0.032$ vs.\
$+0.030$ for real $\dphylo$ at $\lambda_{\mathrm{OU}}{=}10^{-4}$), so
we do not read it as a signature of the taxonomic kernel.

\paragraph{Phylogenetic signal on the Jetz tree.}
\Cref{tab:sup_phylo_signal} reports Mantel $r$ and $K_{\mathrm{mult}}$
on the per-species mean codes against the Jetz tree, together with
the spread of the species means.

\begin{table}[h]
\centering
\small
\caption{\textbf{Phylogenetic signal of the per-species mean shape
codes on the Jetz~\citep{jetz2012} tree.} $K_{\mathrm{mult}}$:
multivariate Blomberg's $K$~\citep{blomberg2003,adams2014} ($1$ =
Brownian expectation, $0$ = no phylogenetic signal); Mantel $r$:
correlation between pairwise $\bbeta$ distance and patristic distance
over the $136$ species pairs; $\sigma_{\bar\beta}$: mean per-dimension
s.d.\ of the per-species means. Permutation $p$ in parentheses ($999$ label shuffles).}
\label{tab:sup_phylo_signal}
\begin{tabular}{lccc}
\toprule
Variant & $K_{\mathrm{mult}}$ ($p$) & Mantel $r$ ($p$) & $\sigma_{\bar\beta}$ \\
\midrule
AniMer+~\citep{animerplus2025} released checkpoint & $0.583$ ($.026$) & $+0.360$ ($.008$) & $0.169$ \\
Cosine LR (no OU)                                   & $0.581$ ($.034$) & $+0.358$ ($.006$) & $0.170$ \\
Cosine LR + OU $\lambda{=}10^{-5}$ (\textbf{ours}, seed 0) & $0.577$ ($.037$) & $+0.351$ ($.007$) & $0.144$ \\
Cosine LR + OU $\lambda{=}10^{-5}$, permuted $\dphylo$ & $0.577$ ($.036$) & $+0.351$ ($.016$) & $0.144$ \\
\midrule
Constant LR + OU $\lambda{=}10^{-4}$, real $\dphylo$ & $0.618$ ($.023$) & $+0.317$ ($.032$) & $0.105$ \\
Constant LR + OU $\lambda{=}10^{-4}$, permuted $\dphylo$ & $0.627$ ($.019$) & $+0.328$ ($.028$) & $0.105$ \\
\bottomrule
\end{tabular}
\end{table}

At the headline strength both statistics coincide with the no-OU
recipe and with the released checkpoint, so the prior neither adds
nor removes phylogenetic signal; what moves is the spread of the
species means ($0.170\to0.144$), the tightening visible in
\cref{fig:phylo_calibration}. At $\lambda_{\mathrm{OU}}=10^{-4}$ the
species means collapse further ($\sigma_{\bar\beta}=0.105$) while
both statistics stay significant against the permutation null
(Mantel $r$ $0.32$, $K_{\mathrm{mult}}$ $0.62$); real and
permuted $\dphylo$ remain closely matched at both strengths.

\textit{Real vs.\ permuted matched at the strongest regularization
strength.}
Under the strongest OU regularization in \cref{tab:sup_phylo_signal}, the real- and
permuted-taxonomy variants are \emph{closely matched} on this
diagnostic ($K_{\mathrm{mult}}$ $0.618$ vs.\ $0.627$, Mantel $r$
$+0.317$ vs.\ $+0.328$, $\Delta\rho_{\text{Jetz}} = -0.009$). Both
pull $\bbeta$ codes into similar continuous-tree geometry; the
CowBird-PCK gap of $+1.76$ pp between them
(\cref{fig:permutation_null}) lives elsewhere in the model's
representations, plausibly in the encoder weights themselves rather
than in the regression head's output codes. The $(\alpha,\sigma)$
trajectory diagnostic (\cref{fig:sup_d5_trajectory}) and the
Procrustes-aligned~\citep{kanazawa2018hmr} vertex-error scale (\cref{tab:sup_vertex_errors})
show the same near-identity on independent metrics.

\paragraph{Local cluster purity (k-NN).}
\label{sec:sup:calibration:purity}
The global statistics above look at the whole code distribution at once; here we zoom in and ask a simpler local question --- if you pick a $\bbeta$ code and look at its ten nearest neighbors in code space, how many of them belong to the same bird family?

For each predicted
$\bbeta$ code on the CtrlAVES3D train split, we compute the fraction
of its $k$-NN ($k{=}10$) that share its Linnaean family (or order)
label. We use a random subsample of $1{,}500$ codes and take
distances in the raw $15$-dimensional code space. This $k$-NN purity is a local-neighborhood-sensitive
diagnostic standard in latent-space embedding evaluation.

\begin{table}[h]
\centering
\small
\caption{\textbf{$\bbeta$-space clustering quality by Linnaean~\citep{ioc-bird-list} rank label.}
Higher $k$-NN purity = $\bbeta$-codes of same family/order are
locally more clumped. Stronger OU~\citep{uhlenbeck1930} regimes monotonically increase
purity.}
\label{tab:sup_cluster_purity}
\begin{tabular}{lccc}
\toprule
Variant & $k$-NN purity (family) & $k$-NN purity (order) & Silhouette (family) \\
\midrule
AniMer+~\citep{animerplus2025} retrain @30K (no OU)        & $0.621$ & $0.714$ & $-0.043$ \\
Cosine LR (no OU)                   & $0.610$ & $0.705$ & $-0.042$ \\
Cosine LR + OU $\lambda{=}10^{-5}$  & $0.678$ & $0.755$ & $-0.034$ \\
$\lambda{=}10^{-4}$ + real $\dphylo$ & $\mathbf{0.725}$ & $\mathbf{0.793}$ & $\mathbf{-0.018}$ \\
\bottomrule
\end{tabular}
\end{table}

\paragraph{Reading.}
The OU prior consistently \emph{increases} local cluster purity,
relative to the cosine-LR run without the prior. Family-level purity
rises by $+6.8$ percentage points (pp) at our headline configuration
and by $+11.5$ pp under stronger regularization; order-level purity
rises by $+5.0$ pp and $+8.8$ pp. Silhouette scores are
negative for all variants (the $\bbeta$ codes of different families
overlap substantially), but $k$-NN purity, a more
local, cluster-shape-agnostic measure, isolates the OU-induced
tightening of within-family neighborhoods.

\paragraph{Reconciling Spearman / purity / phylogenetic signal.}
The three measurements probe different aspects of the code geometry:
\begin{itemize}
\item \emph{Global rank correlation} (Spearman $\rho$,
\cref{tab:sup_beta_phylo_correlation}): drops $-0.012$ under OU.
\item \emph{Phylogenetic signal} (Mantel $r$, $K_{\mathrm{mult}}$,
\cref{tab:sup_phylo_signal}): unchanged under OU ($r$
$0.358\to0.351$, $K_{\mathrm{mult}}$ $0.581\to0.577$).
\item \emph{Local cluster purity} ($k$-NN, this table): rises by
$+6.8$ pp at our headline configuration (within-family
neighborhoods are locally tighter).
\end{itemize}
Together: under OU the $\bbeta$ codes have tighter within-family
local structure and slightly looser discrete-rank ordering, while the
phylogenetic signal inherited from pre-training is preserved.

\section{Theoretical refinement: MAP view and structural comparison to SupCon}
\label{sec:sup:theory_refinement}

The main text gives the OU prior a working definition and an
empirical justification. This appendix sharpens the theoretical
reading along three axes: a Bayesian maximum-a-posteriori (MAP)
correspondence
(\cref{sec:sup:theory:map}) that recasts the regularizer as a
tempered Gaussian prior rather than an ad-hoc penalty; a structural
comparison to $\Lcon$~\citep{supcon2020} (\cref{sec:sup:theory:info})
showing OU exposes a $5$-level ordinal where SupCon sees only a
binary indicator; and a clean restatement of the $\alphainf$ limit
(\cref{sec:sup:theory:limit}) that handles the singular-covariance
pitfall the main text glosses over and recovers SupCon as a
degenerate boundary case. Together
they form the structural substrate for the claims in
\cref{sec:method:limit,sec:sup:method_extended}.

Why does this matter empirically? The main text's
real-vs-permuted-$\dphylo$ ablation (+0.67 CowBird AUC, +1.76 PCK;
\cref{fig:permutation_null}) is the empirical lever the theory
predicts. A label permutation preserves the kernel spectrum and the
number of species pairs at each rank distance, but destroys which
species occupy them, i.e.\ the $5$-level
ordinal alignment between species and rank distance. That
ordinal-alignment information is exactly the signal this theory
identifies as $\LOU$'s contribution beyond $\Lcon$. The remainder of
this section makes that signal precise.

\subsection{MAP estimator equivalence}
\label{sec:sup:theory:map}

Is the OU regularizer really a
Bayesian prior in disguise, or just a convenient quadratic penalty?
We show it is the former: minimizing $\mathcal{L}_{\mathrm{base}} +
\lambda \mathcal{L}_{\mathrm{OU}}$ is exactly MAP under a tempered
Gaussian prior with temperature $1/\lambda$. The temperature
interpretation gives $\lambda$ a meaning beyond ``regularization
strength.''

Let $I=(I_1,\dots,I_B)$ be a mini-batch of images with species labels
$s\in\{1,\dots,N\}^B$ ($N{=}17$), and $\bbeta\in\mathbb{R}^{B\times D}$
the predicted AVES~\citep{aves2021} shape codes. The task losses of AniMer+~\citep{animerplus2025} (keypoint $L_1$,
silhouette $L_2$, AVES regression) define an empirical NLL of $I$ given
$\bbeta$ that we treat, up to constants in $\bbeta$, as the
empirical image NLL
\begin{equation}
\mathcal{L}_{\mathrm{base}}(\bbeta;I)
\;=\;-\log p_{\mathrm{img}}(I\mid \bbeta)+C_{\!I},
\label{eq:base_as_nll}
\end{equation}
where $C_{\!I}$ collects $\bbeta$-independent terms. With kernel
$\Sigma_{ij}=\OUcov{i}{j}$ on the $B$-tuple of training species and a
shared mean $\mu\in\mathbb{R}^D$, place the OU prior~\citep{uhlenbeck1930,hansen1997}
\begin{equation}
p(\bbeta\mid\alpha,\sigma,\mu,\dphylo)
\;=\;\prod_{d=1}^{D}\mathcal{N}\!\big(\bbeta_{\cdot,d}\mid\mu\,\mathbf{1},\Sigma\big).
\label{eq:ou_prior}
\end{equation}

\begin{proposition}[MAP correspondence]
\label{prop:map}
For any $\lambda>0$, the minimizer of
$\mathcal{L}_{\mathrm{total}}(\bbeta)
=\mathcal{L}_{\mathrm{base}}(\bbeta;I)+\lambda\LOU(\bbeta)$
is the maximum-a-posteriori estimate of $\bbeta$ under the
\emph{tempered posterior}
\begin{equation}
p_\lambda(\bbeta\mid I)\;\propto\;
   p_{\mathrm{img}}(I\mid\bbeta)\;
   p(\bbeta\mid\alpha,\sigma,\mu,\dphylo)^{\lambda},
\label{eq:tempered}
\end{equation}
i.e., the temperature $1/\lambda$ on the prior. In particular, $\lambda{=}1$
recovers the standard Bayes posterior; $\lambda{<}1$ down-weights
the OU prior (more data-driven), $\lambda{>}1$ sharpens it.
\end{proposition}

\begin{proof}
Take the negative log of \cref{eq:tempered}. By \cref{eq:base_as_nll}
the data term contributes $\mathcal{L}_{\mathrm{base}}+C_I$; by
\cref{eq:lou,eq:ou_prior} the prior contributes
$\lambda(\LOU(\bbeta)+C_\Sigma)$ with $C_\Sigma=\tfrac{DB}{2}\log 2\pi$
(constant in $\bbeta$). Argmin in $\bbeta$ thus equals
$\arg\min_{\bbeta} \mathcal{L}_{\mathrm{base}}+\lambda\LOU$, which is the
total-loss minimizer. The constants $C_I,C_\Sigma$ shift the loss
without affecting the argmin.
\end{proof}

\begin{remark}
Because $(\alpha,\sigma,\mu)$ are themselves learned by SGD on
$\mathcal{L}_{\mathrm{total}}$, the above is a Type-II / empirical-Bayes
joint MAP in $(\bbeta,\alpha,\sigma,\mu)$ rather than fixed-prior MAP;
treating $\Lcon$ as part of $\mathcal{L}_{\mathrm{base}}$ gives a
hierarchical model with two priors (binary-family + continuous-OU),
which our best setting reduces to OU-only by $w_{\mathrm{con}}{=}0$.
\end{remark}

\subsection{Per-pair structural richness compared to SupCon}
\label{sec:sup:theory:info}

What does $\LOU$ see per pair of
training species that $\Lcon$~\citep{supcon2020} does not? SupCon
collapses every pair to a binary same-family / different-family
indicator. $\LOU$ ingests the full Linnaean rank distance, which on
our 17-species matrix takes 5 distinct values. We describe this richer
per-pair signal and show that SupCon arises as
a degenerate limit of $\LOU$ on a binary distance.

\paragraph{Per-pair signal.}
$\Lcon$ uses a binary same-family indicator
$y_{ij}=\mathbf{1}[f_i{=}f_j]\in\{0,1\}$: every pair is collapsed to
``same family'' or ``different family''. By contrast, $\LOU$ ingests
the full Linnaean rank-distance matrix, whose off-diagonal in our
$17$-species set takes values in $\{1,2,4,5,6\}$
(\cref{tab:dtax_distribution}). Each pair therefore carries a
$5$-level ordinal rather than a binary indicator, distinguishing
within-family, same-order cross-family, and three grades of
cross-order separation that the SupCon~\citep{supcon2020} indicator
lumps together.

\paragraph{Limit interpretation.}
$\Lcon$ corresponds structurally to the limit of an OU prior with a degenerate two-level
taxonomy $\dphylo(i,j)\in\{0,\infty\}$, in which cross-family covariances
vanish and within-family blocks are rank-1. In this limit the per-pair
ordinal collapses to a binary indicator. \Cref{sec:sup:theory:limit} makes the regularized
statement precise.

\subsection{Limit analysis revisited}
\label{sec:sup:theory:limit}

The main text
(\cref{sec:method:limit}) writes the $\alphainf$ link to
$\Lcon$~\citep{supcon2020} informally; that informal version glosses
over a singular-covariance subtlety. Here we restate the limit with
a positive-definite jitter $\varepsilon I$ that keeps the matrix
invertible, then show the result is structurally analogous to (but
not literally equal to) SupCon's same-family pull. ``Structurally
analogous'' is the precise word: the two penalties act on the same
direction (within-family residuals) but live in different feature
spaces.

\begin{proposition}[Regularized $\alphainf$ limit]
\label{prop:limit}
Let $\Sigma_\varepsilon=\Sigma+\varepsilon I$ for $\varepsilon{>}0$, and
hold $\sigma^2/(2\alpha)\equiv c{>}0$ constant as $\alpha{\to}\infty$.
Then for the block-taxonomy
$\dphylo(i,j)=0$ if $f_i{=}f_j$ and $\dphylo(i,j){\ge}1$ otherwise:
\begin{equation}
\lim_{\alpha\to\infty}\,\frac{1}{2}(\bbeta_{\cdot,d}{-}\mu)^\top\Sigma_\varepsilon^{-1}(\bbeta_{\cdot,d}{-}\mu)
\;=\;\sum_{f}\frac{1}{2}\sum_{i:f_i=f}\frac{(\beta_{i,d}{-}\bar\beta_{f,d})^{2}}{\varepsilon}
\;+\;O\!\Big(\frac{1}{c}\Big),
\label{eq:limit_form}
\end{equation}
where $\bar\beta_{f,d}=\frac{1}{n_f}\sum_{i:f_i=f}\beta_{i,d}$ is the
empirical family mean. As $\varepsilon\to 0$ along a positive-definite
sequence (e.g.\ $\varepsilon=10^{-6}$, our implementation), the
quadratic form acts as a per-family shrinkage to the family mean with
infinite stiffness on the same-family direction and finite stiffness
on the cross-family directions.
\end{proposition}

\begin{proof}[Proof sketch]
Under the block-taxonomy, $\Sigma$ becomes block-diagonal with
within-family blocks $c\,J_{n_f}$ ($J=\mathbf{1}\mathbf{1}^\top$,
rank $1$) and zero off-block entries (since
$\exp(-\alpha\dphylo(i,j))\to\mathbf{1}[f_i{=}f_j]$). Adding
$\varepsilon I$, the matrix-inversion lemma gives within-family inverse
$\varepsilon^{-1}I_{n_f}-\tfrac{c}{\varepsilon(\varepsilon+c n_f)}J_{n_f}$.
Decomposing $\beta_{i,d}-\mu_d$ into mean (along $\mathbf{1}_{n_f}$)
and residual components: the mean sees stiffness
$1/(\varepsilon{+}c n_f)\to 1/(c n_f)$, while the residual sees
$1/\varepsilon$, diverging as $\varepsilon\to 0$. Cross-family pairs
contribute zero by block-diagonality. Combining over $D$ and dropping
$O(1/c)$ for $c\gg\varepsilon$ yields \cref{eq:limit_form}.
\end{proof}

\begin{remark}
\Cref{eq:limit_form} is the per-pair Gaussian relaxation of the
same-family pull encoded by SupCon: both shrink within-family residuals and leave
cross-family directions unconstrained. The two are not literally
equal, SupCon uses $\ell_2$-normalized features in cosine space
with a $\log\sum\exp$ objective, while \cref{eq:limit_form} is a
Mahalanobis penalty in raw $\bbeta$-space, so we call them
\emph{structurally analogous}. Family-SupCon is recovered, in this
regularized sense, as the degenerate $\alphainf$ limit of an OU
prior on a binary $\{0,\infty\}$ taxonomy: a strict subset of what
$\LOU$ expresses on a $5$-level rank distance.
\end{remark}

\section{Extended method details}
\label{sec:sup:method_extended}

This appendix has two jobs. First, we lay out, in one place, how our
Gaussian prior $\mathcal{L}_{\mathrm{OU}}$ relates to AniMer+'s
family-contrastive loss $\mathcal{L}_{\mathrm{con}}$, so the reader
can see at a glance what extra structure the OU prior carries and
where the two losses formally meet
(\cref{sec:sup:method_extended:loss}). Second, we list the training
and evaluation data, optimizer settings, and precision details
needed to reproduce \cref{tab:main} from the public AniMer+~\citep{animerplus2025}
checkpoint (\cref{sec:sup:method_extended:data}).

\subsection{\texorpdfstring{From $\mathcal{L}_{\mathrm{con}}$ to $\mathcal{L}_{\mathrm{OU}}$ and the $\alpha\to\infty$ limit}{From L\_con to L\_OU and the alpha-to-infinity limit}}
\label{sec:sup:method_extended:loss}

We write the two losses out side by side, because the difference
between them is clearest at the $\alpha\!\to\!\infty$ limit.

The family supervised-contrastive term $\mathcal{L}_{\mathrm{con}}$ used by AniMer+
operates on per-sample features $z_i$ (the \texttt{cls\_feats} head output
in the public implementation) and family labels $f_i$:
\begin{equation}
\mathcal{L}_{\mathrm{con}} = -\sum_{i}\frac{1}{|P(i)|}\sum_{p \in P(i)}
  \log\frac{\exp(z_i\cdot z_p / \tau)}{\sum_{a \neq i}\exp(z_i\cdot z_a / \tau)},
  \quad P(i) = \{j : f_j = f_i, j \neq i\}.
\end{equation}
Its taxonomic input is a binary same-family indicator per pair (same
family or not).

\paragraph{OU likelihood.}
The Ornstein--Uhlenbeck stochastic differential equation
(SDE)~\citep{uhlenbeck1930,hansen1997} along the
Linnaean taxonomy produces a stationary multivariate Gaussian over
tips with covariance in \cref{eq:ou_cov}; treating the $D$ shape
dimensions as independent draws from a shared species covariance
$\Sigma$ yields $\mathcal{L}_{\mathrm{OU}}$ as the negative
log-likelihood (\cref{eq:lou}), reported per-sample. The full MAP
derivation is given in \cref{prop:map}.

\paragraph{The $\alpha\to\infty$ degenerate limit.}
In the degenerate ``block taxonomy'' limit (within-family distance
$0$, cross-family distance $\infty$), the OU covariance $\Sigma$
becomes block-diagonal and the resulting quadratic form pulls
same-family shape codes together while leaving cross-family pairs
uncoupled, the same qualitative effect as
$\mathcal{L}_{\mathrm{con}}$. The regularized statement of this
limit, including the rank-deficient covariance treatment and the
precise sense in which it is a \emph{structural analogy / Gaussian
probabilistic relaxation} rather than a strict equivalence with
SupCon~\citep{supcon2020} / InfoNCE~\citep{oord2018cpc}, is given in
\cref{prop:limit}. The takeaway is that $\mathcal{L}_{\mathrm{con}}$
is the structural analogue of the binary-indicator degenerate limit of
$\mathcal{L}_{\mathrm{OU}}$;
finite $\alpha\!\sim\!1$ exposes the additional graded
cross-family structure that the discrete family indicator throws
away.

\subsection{Data and evaluation}
\label{sec:sup:method_extended:data}

The point of this subsection is to make our experiments
end-to-end reproducible: the reader should know which images we
trained on, which images we tested on, and the optimizer knobs that
fix the headline numbers.

\paragraph{Training data.} CtrlAVES3D~\citep{ctrlaves3d2024} (6\,464
images, 17 CUB~\citep{cub2011} species), the same AVES split
AniMer+ uses in Table~VI. Animal3D (mammals) and CUB pseudo-3D
mat-file splits carry weight $0$ in all our runs, because we lack
redistributable copies and CtrlAVES3D alone suffices for the AVES
branch.

\paragraph{Evaluation data.} We report on three test sets, each
probing a different aspect of the trained model. \textbf{CUB
test} (official test split, 5\,794 images) measures
 seen-species fine localization and
follows the official AniMer+~\citep{animerplus2025} evaluation.
The CUB training split (5\,994 images) serves only as a development
list for selecting $\lambda_{\mathrm{OU}}$
(\cref{sec:sup:lambda_sweep}; \cref{fig:sup_lambda_vs_metric} is
scored on it); all headline and ablation
numbers use the official test split.
\textbf{CowBird test}~\citep{badger2020} (114 evaluation instances)
measures unseen-species generalization: CowBird species are not
present in CtrlAVES3D~\citep{ctrlaves3d2024} training, making this
the hardest OOD benchmark of the three. The AniMer+ eval protocol
draws those $114$ instances from a $240$-image test split, and our
reported metrics use the same $114$. Finally, \textbf{CtrlAVES3D
test} (309 held-out images) measures seen-species 3D reconstruction.
Metrics follow AniMer+ Table~VI: per-dataset PCK-curve AUC and
PCK@0.1.

\paragraph{Training schedule.} Warm-start from the released
AniMer+ checkpoint;%
\footnote{\url{https://github.com/AniMerPlus/AniMerPlus}}
30K steps, batch 4, AdamW~\citep{loshchilov2017decoupled},
$\eta = 3\!\times\!10^{-7}$ on the main group. 30K steps is
$\sim$2.5 hours on one RTX 4090.

\paragraph{Precision.} Training runs entirely in fp32, including the
ViT~\citep{dosovitskiy2021an} forward/backward and the OU Cholesky,
which avoids under-flow of $\log|\Sigma|$ at small $\alpha$.

\section{The taxonomic distance matrix \texorpdfstring{$D_{\mathrm{tax}}$}{D\_tax}}
\label{sec:sup:dtax_matrix}

The OU prior is parameterized by a fixed $17\!\times\!17$ matrix
$D_{\mathrm{tax}}$ that encodes how related any two species in our
training set are. This appendix exists for two reasons. First, a
reader who wants to recreate our experiments needs to know exactly
which species are in the matrix and what each integer entry means.
Second, the choice of a discrete Linnaean rank over a continuous
molecular phylogeny such as Jetz~\citep{jetz2012} needs a fuller
justification than the main text gives. We address both in turn:
\cref{sec:sup:dtax:species,sec:sup:dtax:construction} document the
matrix; \cref{sec:sup:dtax:why_discrete} explains the design choice.

\subsection{The 17 species and the matrix entries}
\label{sec:sup:dtax:species}

\begin{figure}[h]
\centering
\includegraphics[width=0.9\linewidth]{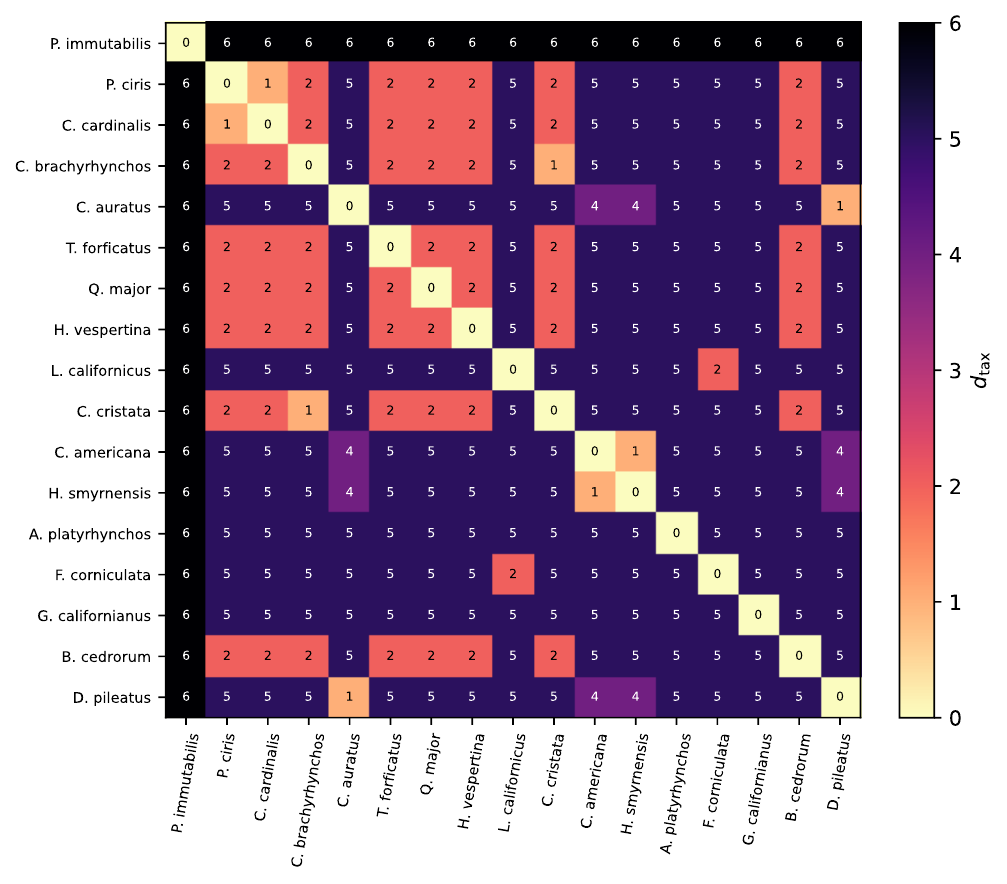}
\caption{\textbf{The full $17\!\times\!17$ taxonomic rank distance matrix $D_{\mathrm{tax}}$ used by the OU~\citep{uhlenbeck1930} prior.}
Diagonal entries are $0$ (same species); off-diagonal entries take
values $\{1, 2, 4, 5, 6\}$, representing same family, same order,
sister-order, cross-order, and outgroup respectively. The species
ordering on the axes matches \cref{tab:species_list}; the dark first
row/column reflect the single-species Procellariiformes outgroup
(\emph{Phoebastria immutabilis}, $i=0$).}
\label{fig:dtax_matrix}
\end{figure}

\Cref{tab:species_list} lists the species and their Linnaean order
and family; the species set is fixed by
CtrlAVES3D~\citep{ctrlaves3d2024} and we did not curate it. The row
index $i$ of each species in $D_{\mathrm{tax}}$ matches the species
ordering in \cref{fig:dtax_matrix}.

\begin{table}[h]
\centering
\small
\caption{\textbf{The 17 CtrlAVES3D species in row-index order in $D_{\mathrm{tax}}$.}
Eight species belong to Passeriformes; the remaining nine span
six non-Passeriformes orders. Species set:
CtrlAVES3D~\citep{ctrlaves3d2024}; binomial nomenclature, family,
and order classification follow the IOC World Bird
List~\citep{ioc-bird-list}.}
\label{tab:species_list}
\begin{tabular}{clll}
\toprule
$i$ & Species (binomial) & Family & Order \\
\midrule
0  & \emph{Phoebastria immutabilis}      & Diomedeidae    & Procellariiformes \\
1  & \emph{Passerina ciris}              & Cardinalidae   & Passeriformes \\
2  & \emph{Cardinalis cardinalis}        & Cardinalidae   & Passeriformes \\
3  & \emph{Corvus brachyrhynchos}        & Corvidae       & Passeriformes \\
4  & \emph{Colaptes auratus}             & Picidae        & Piciformes \\
5  & \emph{Tyrannus forficatus}          & Tyrannidae     & Passeriformes \\
6  & \emph{Quiscalus major}              & Icteridae      & Passeriformes \\
7  & \emph{Hesperiphona vespertina}      & Fringillidae   & Passeriformes \\
8  & \emph{Larus californicus}           & Laridae        & Charadriiformes \\
9  & \emph{Cyanocitta cristata}          & Corvidae       & Passeriformes \\
10 & \emph{Chloroceryle americana}       & Alcedinidae    & Coraciiformes \\
11 & \emph{Halcyon smyrnensis}           & Alcedinidae    & Coraciiformes \\
12 & \emph{Anas platyrhynchos}           & Anatidae       & Anseriformes \\
13 & \emph{Fratercula corniculata}       & Alcidae        & Charadriiformes \\
14 & \emph{Geococcyx californianus}      & Cuculidae      & Cuculiformes \\
15 & \emph{Bombycilla cedrorum}          & Bombycillidae  & Passeriformes \\
16 & \emph{Dryocopus pileatus}           & Picidae        & Piciformes \\
\bottomrule
\end{tabular}
\end{table}

\Cref{fig:dtax_matrix} shows $D_{\mathrm{tax}}$ as a heatmap.
Off-diagonal entries take only five distinct values
$\{1,2,4,5,6\}$ (\cref{tab:dtax_distribution}); each value has a
clear taxonomic interpretation that we explain in
\cref{sec:sup:dtax:construction}.

\begin{table}[h]
\centering
\small
\caption{\textbf{Off-diagonal distribution of $D_{\mathrm{tax}}$.}
Counts are over the $17\times 16 = 272$ ordered off-diagonal pairs;
an unordered species pair contributes $2$ entries (one in each
direction). Diagonal entries (same species, $d=0$) are not
included.}
\label{tab:dtax_distribution}
\begin{tabular}{ccll}
\toprule
$d_{\mathrm{tax}}$ & Count & Pairs & Interpretation \\
\midrule
1 & 8   & 4 unordered  & Same family \\
2 & 54  & 27 unordered & Same order, different family \\
4 & 8   & 4 unordered  & Sister-orders within Telluraves \\
5 & 170 & 85 unordered & Cross-order, non-sister \\
6 & 32  & 16 unordered & Outgroup (Procellariiformes vs.\ all) \\
\midrule
total off-diag & 272 & 136 unordered & \\
\bottomrule
\end{tabular}
\end{table}

\subsection{Construction rules}
\label{sec:sup:dtax:construction}

We assign integer rank distances to each species pair using a
deterministic rule based on Linnaean order and family
(\cref{tab:species_list}). The rule is designed so that ``closer in
the tree'' translates monotonically to ``smaller integer''.

\begin{itemize}
\item $d=0$: same species (diagonal).
\item $d=1$: same family. The four within-family pairs (eight
  ordered entries) are Cardinalidae ($1\!\leftrightarrow\!2$),
  Corvidae ($3\!\leftrightarrow\!9$), Picidae
  ($4\!\leftrightarrow\!16$), and Alcedinidae
  ($10\!\leftrightarrow\!11$).
\item $d=2$: same order, different family. This rank captures the
  Passeriformes-internal pairs not already at rank $1$ (e.g.\
  $1\!\leftrightarrow\!3$, Cardinalidae--Corvidae) and the two
  Charadriiformes families (Laridae--Alcidae,
  $8\!\leftrightarrow\!13$).
\item $d=4$: sister-orders within the Telluraves clade. We treat
  Piciformes and Coraciiformes as sister-orders following the
  whole-genome avian phylogeny~\citep{jarvis2014}, which places both
  within Telluraves with strong support; the four unordered pairs
  are Picidae--Alcedinidae:
  $4\!\leftrightarrow\!10,\ 4\!\leftrightarrow\!11,\
  16\!\leftrightarrow\!10,\ 16\!\leftrightarrow\!11$.
\item $d=5$: cross-order, non-sister. All other cross-order pairs
  among Anseriformes, Charadriiformes, Cuculiformes, and the
  remaining Passeriformes / Piciformes / Coraciiformes pairs that
  are not already at rank $4$. These orders do not have a
  closely-related sister-order within the 17-species subset, so we
  assign them a single rank rather than try to discriminate among
  them at this granularity.
\item $d=6$: outgroup. The oceanic Procellariiformes
  (\emph{Phoebastria immutabilis}, $i=0$) is the only pelagic
  seabird in the set and is assigned the deepest rank as a design
  choice, so its row and column take rank $6$. This is an ordinal
  placement, not a claim about the root of the avian tree: in the
  whole-genome phylogeny~\citep{jarvis2014} the deepest split among
  our species separates Anseriformes (Galloanseres) from the
  remaining Neoaves.
\end{itemize}

The cross-order rank values $\{4,5,6\}$ are ordinal (``sister-order
$<$ generic cross-order $<$ deep outgroup'') and the OU kernel's
length-scale parameter $\alpha$ (\cref{eq:ou_cov}) absorbs the
ordinal-to-cardinal conversion: the learned $\alpha$ converges to
$\sim\!1.0$ across our runs (\cref{sec:sup:trajectories}), so
cross-order similarity decays smoothly and monotonically with rank,
and a uniform rescaling of $d$ is equivalent (up to a redefinition
of $\alpha$) to leaving the ordinal values as written.

\subsection{Why discrete Linnaean rank rather than continuous Jetz distance}
\label{sec:sup:dtax:why_discrete}

A natural alternative is to use a continuous molecular phylogeny
such as the Jetz~\citep{jetz2012} patristic distance. We retain a
Jetz patristic distance matrix for diagnostic use in
\cref{sec:sup:calibration}: the posterior mean across $100$
Hackett backbone trees, in millions of years, with range
$[18.6, 201.2]$\,My. We do not train against it directly.
The reason is the kernel-conditioning argument of
\cref{sec:method:prelim}, which we restate here.

On the 17-species CtrlAVES3D subset, the rank-correlation between
$D_{\mathrm{tax}}$ and the Jetz patristic matrix is high (Pearson
$0.85$, Spearman $0.76$), so the discrete rank distance preserves
most of the biological signal of a continuous molecular kernel.
What it does not preserve is the dynamic range. Raw Jetz distances
span $\sim$$200$\,My, which would force the OU length-scale
$\alpha$ near $10^{-2}$ to express any cross-family correlation;
the discrete Linnaean ordinals $\{1,2,4,5,6\}$ instead lie on the
same order as the converged $\alpha\!\sim\!1$, yielding a
well-conditioned $\Sigma$ at the operating point the optimizer
selects on its own. Discrete Linnaean rank therefore gives us most
of the biological information of the continuous tree while keeping
the kernel in the regime where the OU prior is doing real work.

\section{Extended experimental results}
\label{sec:sup:extended_experiments}

This appendix supports the headline numbers in \cref{tab:main} from
several angles, so the reader can check that the gain is not specific
to a single choice. We probe robustness across repeated runs,
sensitivity to the OU regularization strength, the role of the OU-LR
multiplier,
 the behavior of the real-vs-permuted contrast outside the
training taxonomy, and the training dynamics of the OU scalars
themselves. Together these views are intended to show that the
unseen-species improvement is stable across the design knobs we vary.

\subsection{Robustness across multiple runs}

To check that the headline numbers in \cref{tab:main} do not
depend on a single run, we repeat the headline
operating point five times with different random seeds
(the only thing that changes across runs) and report all five results
in \cref{tab:sup_seeds}.

\begin{table}[h]
\centering
\small
\caption{\textbf{Per-trial results at the headline operating point.}
Full hyperparameter setup in \cref{sec:sup:training_impl}. Each row
is one independent rerun (training seed in parentheses); \cref{tab:main} reports
the Mean $\pm$ s.d.\ row. CUB is scored on the official
test split.}
\label{tab:sup_seeds}
\resizebox{\linewidth}{!}{%
\begin{tabular}{lccccc}
\toprule
Run & CUB~\citep{cub2011} AUC & CUB~\citep{cub2011} PCK@0.1 & CowBird~\citep{badger2020} AUC & CowBird~\citep{badger2020} PCK@0.1 & CtrlAVES3D~\citep{ctrlaves3d2024} AUC \\
\midrule
Trial 1 (seed 0)   & 89.31 & 57.91 & 60.91 & 19.26 & 93.04 \\
Trial 2 (seed 7)   & 89.29 & 57.84 & 60.71 & 18.88 & 93.06 \\
Trial 3 (seed 21)  & 89.31 & 57.94 & 60.88 & 18.59 & 93.06 \\
Trial 4 (seed 42)  & 89.27 & 57.71 & 60.80 & 18.64 & 93.05 \\
Trial 5 (seed 123) & 89.30 & 57.82 & 60.70 & 18.92 & 93.05 \\
\midrule
Mean $\pm$ s.d.    & $89.30 \pm 0.02$ & $57.84 \pm 0.09$ & $60.80 \pm 0.10$ & $18.86 \pm 0.27$ & $93.05 \pm 0.01$ \\
\midrule
AniMer+ paper~\citep{animerplus2025} & 89.50 & 59.40 & 59.30 & 16.40 & 93.00 \\
\bottomrule
\end{tabular}%
}
\end{table}

\paragraph{All five runs clear the AniMer+ paper baseline on the
unseen-species benchmark.} The five trials sit in tight bands (CowBird AUC
$60.70$--$60.91$, CowBird PCK $18.59$--$19.26$,
CtrlAVES3D AUC $93.04$--$93.06$, CUB AUC $89.27$--$89.31$, CUB PCK
$57.71$--$57.94$), and the CowBird and CtrlAVES3D bands lie above
the corresponding AniMer+ Table~VI numbers. The unseen-species gain
is therefore not specific to a single run but repeats across
independent reruns at the same operating point. \Cref{tab:main} reports the mean $\pm$ sample standard deviation of these five runs; this table provides the underlying per-run values
at the same operating point.

\subsection{\texorpdfstring{Sensitivity to $\lambda_{\mathrm{OU}}$ and the seen-vs-unseen trade-off}{Lambda sensitivity}}
\label{sec:sup:lambda_sweep}

We next examine how sensitive the headline numbers are to the
strength of the OU regularizer, $\lambda_{\mathrm{OU}}$, and how that
strength trades seen-species against unseen-species accuracy.

\paragraph{Seen vs.\ unseen species.}
We call a species \emph{seen} if the model was trained on it at either
of its two training stages: AniMer+~\citep{animerplus2025}
pre-training, which includes the CUB~\citep{cub2011} training split
and therefore all $200$ CUB species, or our fine-tuning on
CtrlAVES3D~\citep{ctrlaves3d2024}, which covers $17$ of those
species. Both seen-species benchmarks use held-out images: the
CtrlAVES3D test split contains the $17$ fine-tuning species, and the
CUB test split contains all $200$ CUB species, $183$ of which are
outside our fine-tuning taxonomy (analyzed separately in
\cref{sec:sup:unseen_cub}). A species is \emph{unseen} if it appears
at neither stage; the unseen-species benchmark is
CowBird~\citep{badger2020}.

\begin{figure}[h]
\centering
\includegraphics[width=0.78\linewidth]{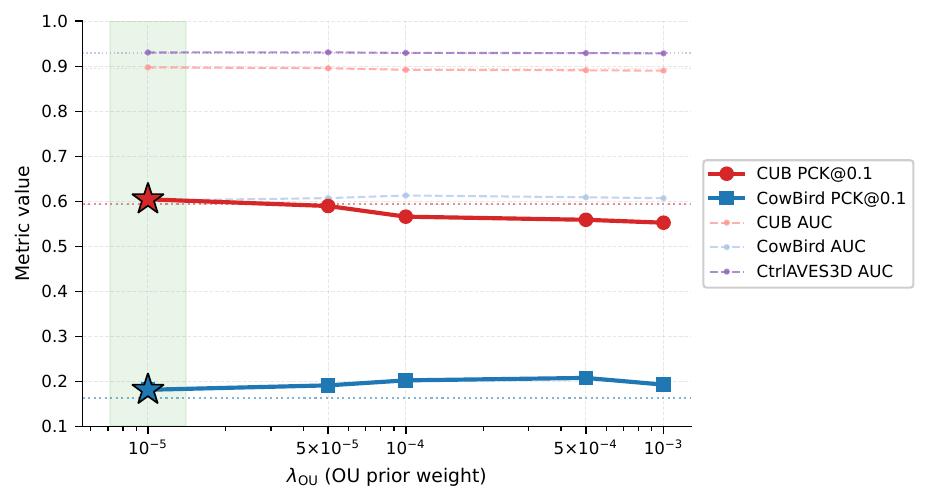}
\caption{\textbf{Evaluation metrics across the swept OU~\citep{uhlenbeck1930} regularization strength $\lambda_{\mathrm{OU}}$.}
Solid lines are PCK@0.1 on the seen-species CUB~\citep{cub2011} and the
unseen-species CowBird~\citep{badger2020}; dashed lines are AUC. Dotted
lines are the AniMer+ paper values; the shaded region and stars mark
$\lambda_{\mathrm{OU}}=10^{-5}$, the value we adopt. All points are constant-LR runs. The sweep is scored on
the development list of \cref{sec:sup:method_extended}. The figure illustrates the
seen-vs-unseen trade-off: seen-species accuracy regresses smoothly
with $\lambda_{\mathrm{OU}}$ while unseen-species accuracy peaks at
intermediate values.}
\label{fig:sup_lambda_vs_metric}
\end{figure}

\Cref{fig:sup_lambda_vs_metric} traces how the proposed prior trades
seen against unseen accuracy as $\lambda_{\mathrm{OU}}$ varies. The
seen-species CUB-PCK regresses smoothly as $\lambda_{\mathrm{OU}}$
grows --- the cost of over-smoothing fine localization on species the
model has already seen --- while the unseen-species CowBird-PCK rises
and peaks at intermediate values, the unseen-species benefit. On the
development list of \cref{sec:sup:method_extended}, the smallest
$\lambda_{\mathrm{OU}}$ that still produces a
measurable taxonomic effect (family $k$-NN purity $0.61\to0.68$,
\cref{sec:sup:calibration}) is
$\lambda_{\mathrm{OU}} = 10^{-5}$, which we adopt as the headline
operating point.

\subsection{Real vs.\ permuted outside the training taxonomy}
\label{sec:sup:unseen_cub}

CowBird covers a single unseen species. We therefore also compare
the real- and permuted-$D_{\mathrm{tax}}$ checkpoints of
\cref{fig:permutation_null} on the $183$ CUB~\citep{cub2011}
species absent from the $17$-species fine-tuning taxonomy (official
test split, $5{,}289$ images; \cref{tab:sup_unseen_cub}). Because images of
one species are correlated, the confidence intervals resample species
rather than images ($10{,}000$ draws), and we also report the
fraction of species on which the real matrix is better. Over all
$183$ species the real matrix is ahead on both metrics with
species-clustered intervals excluding zero; the two order groups show
similar point estimates, with the smaller non-Passeriformes subset
not reaching significance on PCK.

\begin{table}[h]
\centering
\small
\caption{\textbf{Real vs.\ permuted $D_{\mathrm{tax}}$ on the $183$
CUB species outside the fine-tuning taxonomy} (official test
split; same checkpoints as \cref{fig:permutation_null}). $\Delta$ =
real $-$ permuted in percentage points; $95\%$ intervals from a
species-clustered bootstrap; ``real $>$ perm.'' counts species whose
mean is strictly higher under the real matrix.}
\label{tab:sup_unseen_cub}
\resizebox{\linewidth}{!}{%
\begin{tabular}{lcccccccc}
\toprule
Group & Images & Species & $\Delta$ PCK@0.1 & $95\%$ CI & real $>$ perm. & $\Delta$ AUC & $95\%$ CI & real $>$ perm. \\
\midrule
All $183$ species & $5{,}289$ & $183$ & $+0.28$ & $[+0.10, +0.46]$ & $98/183$ & $+0.05$ & $[+0.04, +0.07]$ & $128/183$ \\
Passeriformes     & $3{,}679$ & $125$ & $+0.29$ & $[+0.09, +0.49]$ & $66/125$ & $+0.05$ & $[+0.04, +0.07]$ & $90/125$ \\
Non-Passeriformes & $1{,}610$ & $58$  & $+0.27$ & $[-0.10, +0.62]$ & $32/58$  & $+0.06$ & $[+0.01, +0.10]$ & $38/58$ \\
\bottomrule
\end{tabular}%
}
\end{table}

\subsection{Full OU-LR multiplier ablation}

The OU-LR multiplier $k$ sets the learning rate of a separate AdamW
parameter group holding $\alpha$, $\sigma$ and $\mu$ to $k$ times the
backbone rate, as in discriminative fine-tuning~\citep{howard2018ulmfit}.
We sweep $k$ to separate the effect of the prior's shape at
initialization from that of learning $\alpha$ and $\sigma$, and find
that evaluation is nearly unchanged for $k\in\{1,10,30\}$. The prior
therefore acts through its shape, not through fitted scalars.

\begin{table}[h]
\centering
\small
\caption{\textbf{Effect of OU-LR multiplier $k$ on eval at $\lambda_{\mathrm{OU}} =
10^{-4}$.}}
\label{tab:sup_lrmult}
\begin{tabular}{lcccc}
\toprule
$k$ & CowBird~\citep{badger2020} PCK@0.1 & $\alpha_{\mathrm{end}}$
    & $\sigma_{\mathrm{end}}$ & $\alpha,\sigma$ moving? \\
\midrule
1   & 18.99 & 1.001 & 1.000 & No (frozen at init) \\
10  & 18.93 & 1.005 & 0.996 & Barely ($\leq 0.5\%$) \\
30  & 19.09 & 1.013 & 0.988 & Slightly ($\sim$1\%) \\
100 & 20.25 & 1.038 & 0.963 & Yes ($\sim$3.7\%) \\
\bottomrule
\end{tabular}
\end{table}

\paragraph{The key observation: eval is nearly $k$-invariant at
$\lambda = 10^{-4}$.} CowBird PCK differs by $\leq 0.16$\,pp across
$k\in\{1,10,30\}$, despite $\alpha$ moving from
frozen at $k=1$ to $\sim$1\% at $k=30$. Only at $k=100$ does eval
meaningfully shift ($+1$\,pp CowBird PCK).

\paragraph{Interpretation.} The OU prior's eval contribution at fixed
$\lambda_{\mathrm{OU}}$ comes primarily from the \emph{regularization
shape} imposed by $\Sigma$ at initial values $(\alpha_0,\sigma_0) =
(1,1)$, \emph{not} from learned updates. The data-fit updates at $k=100$
produce a secondary effect on CowBird PCK. The headline operating
point uses $k=100$
to keep the theoretical interpretation testable, but the eval gain is
approximately whatever the OU regularization shape alone would give at
the frozen initialization $(\alpha,\sigma)=(1,1)$. This decoupling is independent evidence
against ``our result is an $(\alpha,\sigma)$-fitting artifact.''

\paragraph{$k = 300$ and beyond.} At $\lambda_{\mathrm{OU}} =
5\!\cdot\!10^{-4}$, $k=300$ and $k=1000$ both train to 30K steps
without divergence, and evaluation is unchanged from $k=100$, with
CowBird-PCK $19.57$ for all three. At $k=1000$ the scalars $\alpha$ and
$\sigma$ oscillate during training but stay finite. We use $k=100$ throughout the main
results for consistency.

\subsection{Training dynamics: OU scalar trajectories}
\label{sec:sup:trajectories}

\begin{figure}[h]
\centering
\includegraphics[width=\linewidth]{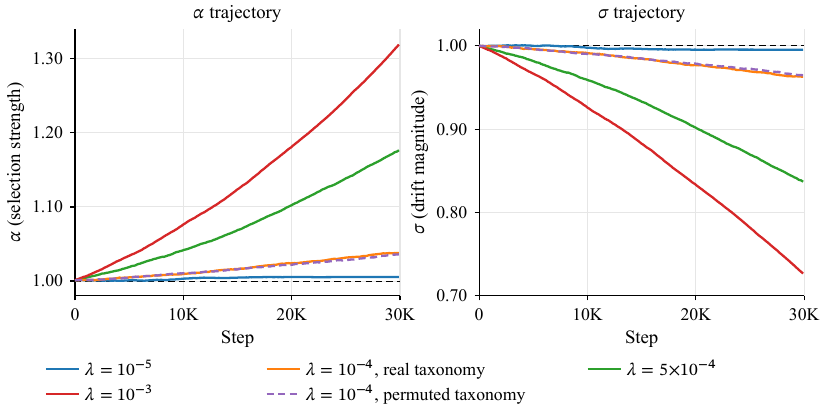}
\caption{\textbf{OU~\citep{uhlenbeck1930} hyperparameter learning trajectories across $\lambda_{\mathrm{OU}}$ values and real vs.\ permuted taxonomies.}
All runs train for 30K steps with $k=100$. Both scalars start at $1$, marked by the dashed line. $\alpha$ rises and $\sigma$ falls in every run, and the change grows with
$\lambda_{\mathrm{OU}}$. The $\lambda_{\mathrm{OU}}=10^{-5}$ run uses a cosine LR and the others a constant LR. The real and permuted taxonomy variants
give nearly identical trajectories, showing that the motion of $\alpha$ and $\sigma$
is driven by the OU kernel \emph{shape} rather than tree
\emph{content}.}
\label{fig:sup_alpha_sigma}
\end{figure}

We finally track the OU~\citep{uhlenbeck1930,hansen1997} scalars
$(\alpha,\sigma)$ during training. \Cref{fig:sup_alpha_sigma} shows
that the direction is the same in every run ($\alpha$ up, $\sigma$
down) and that the magnitude scales smoothly with
$\lambda_{\mathrm{OU}}$, ranging from $\sim$$0.5\%$ at the headline
$\lambda_{\mathrm{OU}}=10^{-5}$ (cosine LR; $\sim$$1.7\%$ for the
constant-LR run at the same $\lambda_{\mathrm{OU}}$) to $\sim$$32\%$ at
$\lambda_{\mathrm{OU}}=10^{-3}$ (the $\lambda_{\mathrm{OU}}=10^{-4}$
endpoints match the $(\alpha_{\mathrm{end}},\sigma_{\mathrm{end}})$
columns of \cref{tab:sup_lrmult} at $k\!=\!100$). Because the real and
permuted taxonomies move the scalars alike, this motion is not evidence
that the data prefer the tree.

\begin{figure}[h]
\centering
\includegraphics[width=\linewidth]{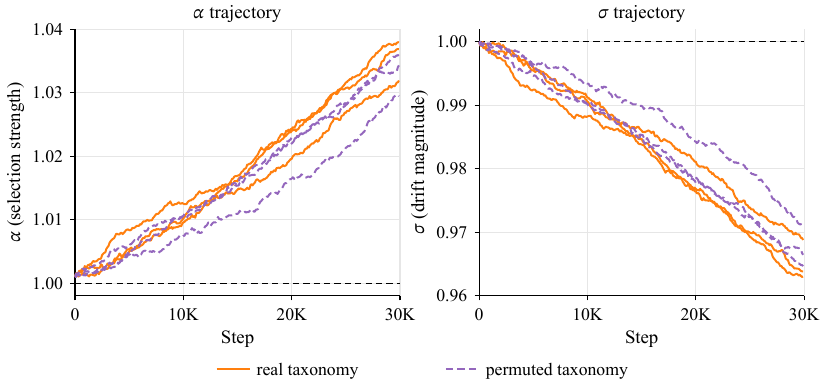}
\caption{\textbf{$\alpha$ and $\sigma$ over 30K steps for the real and permuted $D_{\mathrm{tax}}$ at $\lambda_{\mathrm{OU}} = 10^{-4}$.}
All runs use a constant LR, and each taxonomy has three runs with seeds $0$, $42$ and $123$.
Both scalars start at $1$, marked by the dashed line. The real and permuted curves stay close
throughout training, and the gap between them is on average smaller than the spread across seeds.
\Cref{sec:permutation_ablation} gives the complementary final-eval evidence.}
\label{fig:sup_d5_trajectory}
\end{figure}

The real-$\dphylo$ and permuted-$\dphylo$ runs at the matched
permutation-ablation operating point overlay closely throughout
training (\cref{fig:sup_d5_trajectory}), and for seed $0$ end at close values of
$\alpha$ ($1.038$ vs.\ $1.036$) and $\sigma$ ($0.963$ vs.\ $0.965$).
Scalar motion is therefore driven by the kernel \emph{shape} (matrix
spectrum), not by the biological semantics of the tree; the CowBird
AUC/PCK gap reported in \cref{sec:permutation_ablation} is not
explained by $(\alpha,\sigma)$ endpoint differences. The
$\bbeta$-space calibration diagnostics likewise do not separate the
two runs (\cref{sec:sup:calibration}), and their vertex-error gap is
small (\cref{tab:sup_vertex_errors}).

\section{Vertex-level errors on CtrlAVES3D}
\label{sec:sup:vertex_errors}

In this appendix we report 3D Procrustes-aligned~\citep{kanazawa2018hmr} per-joint and per-vertex errors (pa-MPJPE, pa-MPVPE) on CtrlAVES3D~\citep{ctrlaves3d2024} so the reader can see whether the 2D-keypoint gain in \cref{tab:main} shows up in the 3D mesh, and as a complementary axis of evidence on the real-vs-permuted comparison. The AniMer+~\citep{animerplus2025} paper reports these in its Table~VI (AniMer+: pa-MPJPE $65.6$, pa-MPVPE $70.9$ on CtrlAVES3D), but our re-implemented evaluation uses the canonical AVES template units (roughly $10\times$ that scale), so we benchmark our reimplementation against itself across variants; the use is relative.

\begin{table}[h]
\centering
\small
\caption{\textbf{Procrustes-aligned~\citep{kanazawa2018hmr} 3D errors on the CtrlAVES3D~\citep{ctrlaves3d2024} test split (309 images, 17 species).}
pa-MPJPE = mean per-joint position error after Procrustes alignment;
pa-MPVPE = mean per-vertex position error. Both in canonical AVES
template units (see Caveat). Ours is the seed-$0$ run (Trial~1 of
\cref{tab:sup_seeds}).}
\label{tab:sup_vertex_errors}
\begin{tabular}{lcc}
\toprule
Variant & pa-MPJPE~\citep{kanazawa2018hmr} $\downarrow$ & pa-MPVPE $\downarrow$ \\
\midrule
AniMer+~\citep{animerplus2025} pretrained checkpoint & $655.61$ & $708.97$ \\
AniMer+~\citep{animerplus2025} \textbf{+ OU (ours)} & $\mathbf{649.89}$ & $\mathbf{699.01}$ \\
\midrule
$\lambda{=}10^{-4}$ + real $\dphylo$       & $650.91$ & $697.98$ \\
$\lambda{=}10^{-4}$ + perm $\dphylo$    & $652.15$ & $699.31$ \\
$\lambda{=}5\!\cdot\!10^{-4}$              & $653.74$ & $704.56$ \\
$\lambda{=}10^{-3}$                          & $651.20$ & $701.56$ \\
\bottomrule
\end{tabular}
\end{table}

\paragraph{Reading.}
Our method improves over the AniMer+ pretrained checkpoint by
$\Delta\mathrm{pa\text{-}MPJPE}{=}{-}5.72$ and
$\Delta\mathrm{pa\text{-}MPVPE}{=}{-}9.96$ (canonical units). The vertex-error
scale shows the same direction as the CowBird-PCK ablation in
\cref{fig:permutation_null}: real $\dphylo$
($\mathrm{pa\text{-}MPVPE}{=}697.98$) is slightly better than the
permuted control ($699.31$, $\Delta{=}{-}1.33$). The vertex-error
gap is small relative to the $\sim$$700$-unit scale, consistent with the OU prior's mechanism being more visible
on sparse 2D keypoint metrics than on densely-supervised 3D vertex
positions. The $(\alpha,\sigma)$ trajectory diagnostic
(\cref{fig:sup_d5_trajectory}) and the phylogenetic-signal diagnostic
(\cref{tab:sup_phylo_signal}) do not separate the two variants, so
this small vertex-error difference is the only one of the three
diagnostics that favors the real matrix.

\paragraph{Caveat.}
These numbers come from our re-implementation of the eval pipeline,
not from the official AniMer+ Table~VI, whose CtrlAVES3D values
($65.6$ / $70.9$) and the AniMer~\citep{animer2025} Animal3D
pa-MPJPE ($80.4$\,mm) are on a different unit scale from the
canonical AVES template units used here.
\Cref{tab:sup_vertex_errors} should therefore be read as
\emph{relative} comparisons among our variants on the same eval
pipeline, not as absolute numbers comparable to AniMer's mammal
table.

\section{CowBird OOD per-image error distribution}
\label{sec:sup:cowbird_qualitative}

The headline CowBird~\citep{badger2020} improvement in \cref{tab:main}
($+2.46$ pp PCK over AniMer+~\citep{animerplus2025}) is a \emph{bulk}
statistic. In this appendix we check from a per-image view that it is
broadly distributed across the 114 test images, not driven by a few
outliers.

\paragraph{Setup.}
We run the AniMer+ retrain @30K baseline (no OU) and our headline
checkpoint on all $n{=}114$ CowBird test images, computing per-image
keypoint $L_2$ errors on the visible-mask subset.
\Cref{tab:sup_perimage} instead compares against the released
AniMer+ checkpoint.

\begin{figure}[h]
\centering
\includegraphics[width=\linewidth]{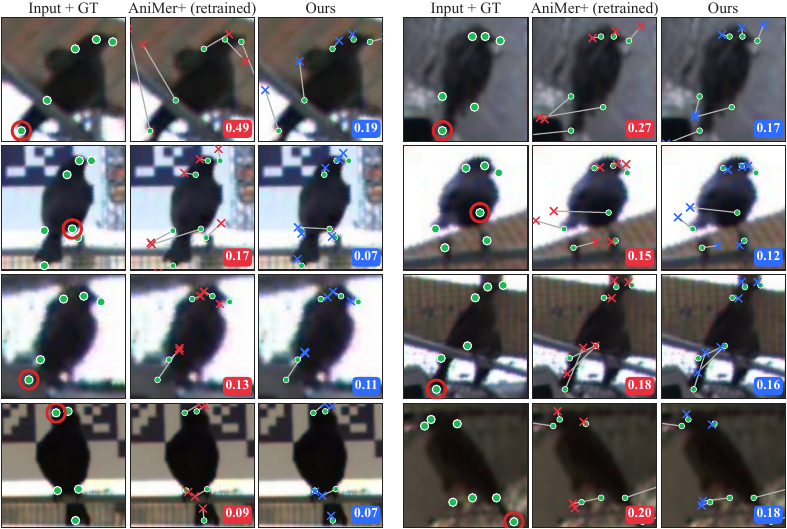}
\caption{\textbf{Eight CowBird~\citep{badger2020} test cases with the largest per-image mean-error reduction under the OU~\citep{uhlenbeck1930} fine-tune.}
The convention follows \cref{fig:qualitative}, with a first panel showing
the input and the ground-truth keypoints in green. The retrained
AniMer+~\citep{animerplus2025} baseline has red crosses and ours has blue
crosses. Gray lines join each prediction to its ground-truth keypoint, and
the red circle marks the keypoint the baseline misplaces most. The
badge is the per-image mean keypoint error over the visible keypoints
in normalized crop units, so lower is better. Best viewed in
electronic format.}
\label{fig:sup_cowbird_top_panel}
\end{figure}

\begin{figure}[h]
\centering
\includegraphics[width=0.7\linewidth]{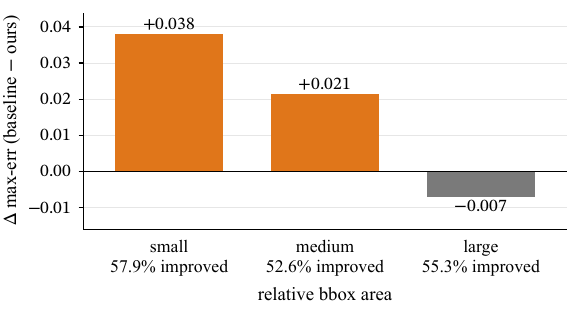}
\caption{\textbf{Per-image $\Delta$ max-err on CowBird~\citep{badger2020} by relative bbox area.}
The 114 test images are split into three terciles of 38 images each. $\Delta$ is baseline minus ours,
so a positive bar means ours is better, and the label under each bar gives the share of images improved.
The improvement is largest on small birds at $+0.038$, smaller on medium birds at $+0.021$, and
slightly negative on large birds at $-0.007$.}
\label{fig:sup_cowbird_bbox_area}
\end{figure}

\paragraph{Where the gain concentrates: small birds.}
\Cref{fig:sup_cowbird_top_panel} shows eight representative cases
with the largest reduction in per-image mean keypoint error from
baseline to ours, among images with at least five visible keypoints
on which ours ends below $0.2$ (normalized crop units). The selected
cases span varied poses and backgrounds, with no obvious pose bias.
Splitting the 114 test images into terciles by relative bbox area
(\cref{tab:sup_cowbird_bbox_area,fig:sup_cowbird_bbox_area})
sharpens the picture: the per-image improvement is largest on small
birds, $+0.038$, smaller on medium birds, $+0.021$, and slightly negative
on large birds, $-0.007$. The
interpretation aligns with the prior's design intent: when
image-level cues are weak (small birds, fewer pixels covering the
body), the species-level shape prior carries proportionally more
useful information; when the bird fills the frame, the per-image
regression head of AniMer+ already has abundant signal and
species-level priors matter less.

\begin{table}[h]
\centering
\small
\caption{\textbf{CowBird~\citep{badger2020} per-image $\Delta$ max-err (baseline $-$ ours) split by bbox area tercile.}
Small birds (small image footprint)
receive the largest OU~\citep{uhlenbeck1930} benefit, medium birds a
smaller one, and large birds a slightly negative change on average.
Max-err is the largest keypoint error in an image.}
\label{tab:sup_cowbird_bbox_area}
\begin{tabular}{lcccc}
\toprule
Bbox-area tercile & $n$ & Mean baseline max-err & Mean ours max-err & $\Delta$ (improvement rate) \\
\midrule
Small (lowest tercile)  & $38$ & $0.435$ & $0.396$ & $\mathbf{+0.038}$ ($57.9\%$ improved) \\
Medium (mid tercile)    & $38$ & $0.503$ & $0.482$ & $+0.021$ ($52.6\%$ improved) \\
Large (highest tercile) & $38$ & $0.620$ & $0.628$ & $-0.007$ ($55.3\%$ improved) \\
\bottomrule
\end{tabular}
\end{table}

\paragraph{Caveats.}
(i) The improvement metric here is \emph{relative} per-image error;
the bulk CowBird-PCK gain comes partially from images that were just
above the PCK threshold under baseline and just below under ours,
which the per-image breakdown does not directly highlight.
(ii) Negative cases (baseline better) do exist and are not selected
against in the top-improvement panel; we display only the largest
ours-better gaps as illustrative examples.

\paragraph{Per-image significance.}
\Cref{tab:sup_perimage} reports a paired per-image comparison of the
seed-$0$ headline run (Trial~1 of \cref{tab:sup_seeds}) against the released AniMer+
checkpoint on the $114$ CowBird instances, with a $10{,}000$-sample
paired bootstrap for the $95\%$ confidence interval and a two-sided
Wilcoxon signed-rank test. All three intervals exclude zero, and AUC
improves on $84$ of the $114$ images, so the gain is broadly
distributed rather than driven by a few outliers. The per-seed
dispersion of the same configuration is in \cref{tab:sup_seeds}.

\begin{table}[h]
\centering
\small
\caption{\textbf{Paired per-image comparison on CowBird~\citep{badger2020}} ($114$ instances; seed-$0$ headline run vs.\ the released AniMer+~\citep{animerplus2025} checkpoint, every image weighted equally).}
\label{tab:sup_perimage}
\begin{tabular}{lccc}
\toprule
Metric & Mean $\Delta$ (pp) & $95\%$ CI & Wilcoxon $p$ \\
\midrule
AUC      & $+1.52$ & $[+0.83, +2.34]$ & $1.5\times10^{-8}$ \\
PCK@0.1  & $+3.21$ & $[+1.35, +5.16]$ & $8.8\times10^{-3}$ \\
PCK@0.15 & $+1.99$ & $[+0.46, +3.68]$ & $1.5\times10^{-2}$ \\
\bottomrule
\end{tabular}
\end{table}

\section{Extended qualitative analysis}
\label{sec:sup:qualitative}

This appendix shows further side-by-side cases in which the OU
fine-tune visibly corrects the AniMer+~\citep{animerplus2025}
baseline; the goal is to illustrate \emph{what kind} of error
reduction the headline numbers in \cref{tab:main} are made of.

\begin{figure}[h]
\centering
\includegraphics[width=\linewidth]{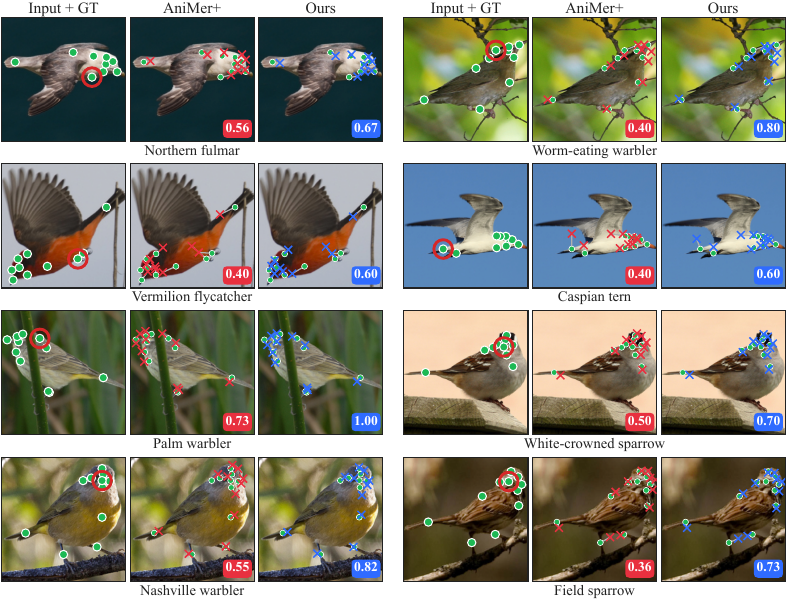}
\caption{\textbf{Eight further CUB~\citep{cub2011} test cases.}
The convention follows \cref{fig:qualitative}, with a first panel showing
the input and the ground-truth keypoints in green. The retrained
AniMer+~\citep{animerplus2025} baseline has red crosses and ours has blue
crosses. The badge is the per-image PCK@0.1, so higher is better. Best viewed in electronic format.}
\label{fig:sup_qualitative}
\end{figure}

\Cref{fig:qualitative} in the main text shows two
CUB~\citep{cub2011} official-test images taken from those with the
largest mean-error reduction among images with at least seven visible
keypoints and our mean keypoint error $\le 0.09$.
\Cref{fig:sup_qualitative} shows eight further cases that pass a fixed
selection rule (baseline worst-keypoint error $\ge 0.12$ in
normalized units, our mean keypoint error $\le 0.10$, mean-error
reduction $\ge 0.015$) among the species outside the $17$-species
fine-tuning taxonomy. Across these eight cases the baseline's largest error falls
on the head (crown, nape, eye, bill base) in five, on the torso
(breast, tail base) in two and on a leg in one. The comparison is
against the AniMer+ baseline retrained with its published recipe for
the same $30{,}000$ steps, so it reflects the full fine-tune rather
than the OU prior in isolation.

\paragraph{Caveat.} These are selected cases, not a statistical
sample: the selection rule picks images on which the baseline fails
and ours does not, so the panels illustrate the kind of correction
the fine-tune makes rather than how often it makes one. Aggregate
behavior is in \cref{tab:main} and, per image, in
\cref{sec:sup:cowbird_qualitative}.

\section{Training implementation details}
\label{sec:sup:training_impl}

This appendix lists the optimizer, numerical, and initialization
choices that determine our headline numbers in \cref{tab:main}. The
warm-start checkpoint, training-data composition, total step count,
batch size, base learning rate, and fp32 precision are listed in
\cref{sec:sup:method_extended}; here we cover only the items that
section does not.

\paragraph{Operating points.}
Headline results (\cref{tab:main,tab:vertex,tab:sup_seeds} and the
qualitative figures) use the cosine LR schedule,
$\lambda_{\mathrm{OU}}=10^{-5}$, $k=100$, $w_{\mathrm{con}}=0$,
$30{,}000$ steps, batch $4$; \cref{tab:main} reports the mean $\pm$ sample
standard deviation over five seeds of this configuration and
\cref{tab:sup_seeds} the per-seed values.
The real-vs-permuted and rank-granularity ablations
(\cref{fig:permutation_null,tab:family_block,tab:sup_lrmult}) use the
stronger-prior setting $\lambda_{\mathrm{OU}}=10^{-4}$ with a
constant LR, chosen so that the taxonomic content of
$D_{\mathrm{tax}}$ is expressed strongly enough for the contrast to
be visible. Both settings share the warm-start checkpoint, data, step
budget and $k$; only $\lambda_{\mathrm{OU}}$ and the schedule differ.
The remaining sweeps (\cref{fig:sup_lambda_vs_metric} and the
multi-variant comparisons in
\cref{tab:sup_beta_phylo_correlation,tab:sup_phylo_signal,tab:sup_cluster_purity,tab:sup_vertex_errors})
state their settings in place.

\subsection{Optimizer and schedule}

We use AdamW~\citep{loshchilov2017decoupled,loshchilov2017sgdr}
with two parameter groups. The base learning rate is
$\eta = 3\!\times\!10^{-7}$ on the main group; the OU group
($\alpha,\sigma,\mu$) lives at $k\eta = 3\!\times\!10^{-5}$ with
$k\!=\!100$. The cosine schedule uses $T_{\max}\!=\!30{,}000$ steps
and $\eta_{\min}\!=\!0.01\eta\!=\!3\!\times\!10^{-9}$. The
$\lambda_{\mathrm{OU}}$ and $k$ axes in
\cref{sec:sup:extended_experiments} jointly identify the operating
point above.

\paragraph{Choice of $k$.}
We use $k\!=\!100$ so that $\alpha$ and $\sigma$ are learned rather
than left at their initial values. AdamW changes a parameter by
roughly the learning rate per step, so at the backbone LR
$\eta = 3\!\times\!10^{-7}$ the OU scalars can move by at most about
$10^{-2}$ over $30$K steps; with $k\!=\!1$ they end within $0.1\%$ of
their initial values (\cref{tab:sup_lrmult}) and the prior acts as a
fixed kernel. Evaluation is nearly unchanged
for $k\in\{1,10,30\}$ and shifts by about one CowBird-PCK point at
$k\!=\!100$ (\cref{tab:sup_lrmult}).

\subsection{Numerical safeguards and initialization}

Training runs in fp32, which avoids the conditioning issues of fp16
on near-singular kernels: the OU covariance assembly, Cholesky
decomposition $\Sigma = LL^{\top}$, and the Mahalanobis quadratic
in \cref{eq:lou} are all computed in fp32. We add a diagonal jitter
$\varepsilon = 10^{-6}$ to $\Sigma$. Without it, two images of the
same species in a mini-batch have $d_{\mathrm{tax}}=0$, which makes
$\Sigma$ singular and the Cholesky fails. Gradient clipping
is at norm $1.0$ on the full parameter set. The OU group has weight
decay $0$: zero weight decay on the OU group prevents AdamW from
driving $\alpha,\sigma$ back toward their softplus-init independently
of the gradient signal, which would interfere with the theoretical
interpretation.

The OU scalars $(\alpha,\sigma)$ use a softplus
parameterization~\citep{dugas2001incorporating} with floor
$\alpha_{\min}=10^{-3}$. The underlying raw scalar is initialized to
$\mathrm{inv\_softplus}(1.0) \approx 0.541$ so that
$\mathrm{softplus}(\mathrm{raw}) = 1.0$ at step zero; thus
$\alpha_0 = \sigma_0 = 1.0$ in the physical (post-softplus)
parameterization. The species-mean $\mu \in \mathbb{R}^{15}$ is
zero-initialized. The
$\mathcal{L}_{\mathrm{con}}$~\citep{supcon2020} weight
$w_{\mathrm{con}}$ used by AniMer+ is set to $0$ in our headline runs,
leaving the OU prior as the sole taxonomy-aware regularizer; setting
$w_{\mathrm{con}}\!=\!5{\times}10^{-4}$ recovers the original AniMer+
configuration.

\subsection{Loss-term integration and permutation control}

To minimize drift from the released AniMer+ codebase we implement
the OU prior as an additive override of AniMer+'s loss computation:
the original task losses are computed unchanged, then
$\mathcal{L}_{\mathrm{OU}}$ is added before the backward pass. This
keeps the parameter ordering identical to the released checkpoint
and guarantees that $\lambda_{\mathrm{OU}}=0$ with $w_{\mathrm{con}}$ left at its
original value recovers the original
training loop to floating-point equality.

For the permutation ablation (\cref{sec:permutation_ablation}) we
permute the species labels on $D_{\mathrm{tax}}$ via a random
permutation $\pi$ applied to both row and column indices. This
preserves the marginal pairwise distance distribution exactly, and
therefore the OU-kernel spectrum, but destroys the biological
correspondence between species labels and taxonomy positions. A
single fixed permutation is used for every
seed and every setting. The selected permutation breaks all four
same-family cross-species relations and reassigns the same-order
relations, so the informative close-kin entries are scrambled.

\subsection{Inference cost}

The OU prior is a training-time loss only: at inference the network
is structurally identical to AniMer+, so no extra parameters,
layers, or kernels are introduced on the forward path. To make this
concrete we measured wall-clock latency and peak GPU memory across
three checkpoints on a single RTX 4090 at batch sizes $\{1,4,8\}$, with
$3$ warm-up iterations and $20$ timed iterations per cell. The
checkpoints are the released AniMer+, our paper-exact retrain at
$30$K steps, and an OU fine-tune at $\lambda_{\mathrm{OU}}=10^{-5}$
with the constant-LR recipe, whose architecture is identical to the
headline run. \Cref{tab:sup_inference_cost} shows the result.

\begin{table}[h]
\centering
\small
\caption{\textbf{Inference-time cost across the three checkpoints.}
Measured on a single RTX 4090 with fp32 forward passes. Latency is
the median over $20$ timed iterations after $3$ warm-up steps;
peak memory and parameter counts are deterministic. The architecture
is unchanged across the three rows, so parameter counts coincide
exactly.}
\label{tab:sup_inference_cost}
\begin{tabular}{lcccc}
\toprule
Variant & ms @ bs$=$1 & ms @ bs$=$8 & Peak mem @ bs$=$8 (MB) & Params (M) \\
\midrule
AniMer+~\citep{animerplus2025} pretrained         & 19.0 & 74.0 & 3160.9 & 785.67 \\
AniMer+ retrain @30K   & 18.9 & 73.9 & 3160.9 & 785.67 \\
\textbf{Ours} & 18.9 & 74.0 & 3160.9 & 785.67 \\
\bottomrule
\end{tabular}
\end{table}

Latency differences are $\leq 0.2$~ms across variants at every batch
size, within the run-to-run jitter: the per-cell p25--p75 spread
is $\approx 0.2$~ms in our measurements. Peak memory and parameter
count are bit-identical by construction. The OU prior therefore
carries no inference-time cost.

\paragraph{Training-time cost of the OU term.}
The OU covariance is assembled over the $B$ instances of the
mini-batch, not over all $N$ species, so one OU step (distance
gather, covariance build, Cholesky factorization and solve) at
$B=4$ takes $0.67$--$0.69$\,ms for $N = 17$, $200$, $2{,}000$ and
$10{,}000$ species (RTX 5090, mean of $100$ iterations after $10$
warm-up steps): the cost is set by $B$, not by $N$. The $N{\times}N$
distance matrix is a fixed lookup table that is never inverted.

\section{Extended discussion: limitations and future work}
\label{sec:sup:future_work}

This appendix expands on the limitations and future directions summarized at the end of \cref{sec:conclusion}, and lays out the three follow-up directions we see as most natural: lifting the pipeline to multi-view 3D reconstruction, replacing the discrete Linnaean rank distance with a continuous molecular phylogeny, and porting the same OU machinery to mammals. Each direction adds exactly one ingredient to the present setup -- multiple views, a continuous tree, or a non-avian dataset -- while leaving the OU prior itself unchanged. We discuss each in turn, together with the obstacles a follow-up paper would need to address.

\subsection{Multi-view extension}

When several synchronized cameras observe the same bird -- the natural setting for full 3D reconstruction in the wild -- the systems-side extension is a multi-view 3D Gaussian
Splatting pipeline~\citep{kerbl2023gaussian} for birds, using
existing multi-view rigs such as 3D-POP~\citep{3dpop2023} (4-view)
with the OU-regularized AniMer+~\citep{animerplus2025} shape codes as template. The OU prior in this
paper is shape-code-level and orthogonal to the multi-view lifting
machinery, so the extension is a clean composition rather than a
re-derivation. Posing it requires re-implementing the AniMer+ lifting
head as a per-Gaussian regressor, which we leave as future work.

\subsection{Scaling the prior to a continuous molecular phylogeny}

The Linnaean rank distance is biologically grounded and admits a
well-conditioned OU kernel, but it is discrete by construction; replacing it with a continuous, time-calibrated tree would let the prior cover thousands more species.
\Cref{sec:method:prelim} argued on conditioning grounds that a naive
substitution of the Jetz patristic matrix~\citep{jetz2012} would
degenerate the kernel because the molecular distances span hundreds
of millions of years and saturate the OU exponential at the
selection-pressure regime our optimizer converges into. A
length-scale-normalized continuous-distance taxonomy
is the natural next step, and would scale the prior to the much
larger species set covered by the Jetz tree. Two options are
$\exp(-\alpha\,d/\bar d)$ over a normalized molecular phylogeny, and
a learned warping that brings the Jetz distance into the
$O(1)$ regime where the converged selection pressure is meaningful. Our $\bbeta$-space
calibration analysis (\cref{sec:sup:calibration}) provides
preliminary empirical support for this direction.

Scaling the species set also changes the memory footprint of the
taxonomy itself. The per-step OU cost is independent of $N$ because
$\Sigma$ is assembled over the $B$ instances in the mini-batch
(\cref{sec:sup:training_impl}), but
the pairwise distance matrix $D_{\mathrm{tax}}$ is $O(N^2)$: at
$N = 10{,}000$ species it occupies about $400$\,MB in fp32, which
would call for compact integer or triangular storage, or on-demand
taxonomy lookups, rather than a dense float matrix.

\subsection{Mammalian analogue}

The OU kernel structure transfers to any taxonomy -- including mammals, the other major taxonomic group already supported by the AniMer+ codebase -- but the kernel
parameters and the choice of distance measure would need to be
re-fit for non-avian groups. The mammalian branch of the AniMer+~\citep{animerplus2025}
training data is inactive in our experiments because we focused on
the avian long-tail problem; a mammalian replication is a natural
companion paper using the same OU machinery on a different
taxonomic substrate.

\end{document}